\documentclass[
  journal=nlp,
  manuscript=article,
  year=XXXX,
  volume=YY,
]{cup-journal}

\usepackage{latexsym}
\usepackage[T1]{fontenc}
\usepackage[nopatch]{microtype}
\usepackage{url}

\usepackage{xcolor}
\usepackage{booktabs}
\usepackage[normalem]{ulem}

\usepackage{newunicodechar}
\newunicodechar{Ł}{\L}
\newunicodechar{ł}{\l}
\newunicodechar{ř}{\v{r}}
\newunicodechar{š}{\v{s}}
\newunicodechar{ć}{\'c}
\newunicodechar{ó}{\'o}
\newunicodechar{á}{\'a}
\newunicodechar{é}{\'e}
\newunicodechar{ı}{\i}
\newunicodechar{́}{\'}
\newunicodechar{ž}{\v{z}}
\newunicodechar{č}{\v{c}}
\newunicodechar{í}{\'i}
\newunicodechar{ñ}{\~n}
\newunicodechar{ü}{\"u}
\newunicodechar{ö}{\"o}
\newunicodechar{ç}{\c{c}}
\newunicodechar{Ç}{\c{C}}
\newunicodechar{î}{\^\i}

\usepackage[backgroundcolor=gray!20,textsize=footnotesize]{todonotes}

\usepackage{makecell}
\usepackage{pgfplots}
\usepackage{hyperref}
\usepackage{multirow}
\usepackage{amsmath}
\usepackage{subcaption}
\usepackage{wrapfig}
\usepackage{catchfilebetweentags}

\title{Smart Bilingual Focused Crawling of Parallel Documents}

\author{Cristian García-Romero}
\affiliation{Dep. de Llenguatges i Sistemes Informàtics, Universitat d'Alacant, Sant Vicent del Raspeig, 03690, Alacant, Spain}
\email[Cristian García-Romero]{cgarcia@dlsi.ua.es}

\author{Miquel Esplà-Gomis}
\affiliation{Dep. de Llenguatges i Sistemes Informàtics, Universitat d'Alacant, Sant Vicent del Raspeig, 03690, Alacant, Spain}

\author{Felipe Sánchez-Martínez}
\affiliation{Dep. de Llenguatges i Sistemes Informàtics, Universitat d'Alacant, Sant Vicent del Raspeig, 03690, Alacant, Spain}


\begin{document}

\begin{abstract}
Crawling parallel texts ---texts that are mutual translations--- from the Internet is usually done following a brute-force approach: documents are massively downloaded in an unguided process, and only a fraction of them end up leading to actual parallel content. In this work we propose a smart crawling method that guides the crawl towards finding parallel content more rapidly. We follow a neural approach that consists in adapting a pre-trained multilingual language model based on the encoder of the Transformer architecture by fine-tuning it for two new tasks: inferring the language of a document from its Uniform Resource Locator (URL), and inferring whether a pair of URLs link to parallel documents. We evaluate both models in isolation and their integration into a crawling tool. The results demonstrate the individual effectiveness of both models, and highlight that their combination enables us to address a practical engineering challenge: the early discovery of parallel content during web crawling in a given language pair. This leads to a reduction in the amount of downloaded documents deemed useless, and yields a greater quantity of parallel documents compared to conventional crawling approaches.

\end{abstract}

\maketitle

\section{Introduction}\label{se:intro}

Human language technologies have experienced unprecedented progress thanks to the advances in artificial intelligence and the availability of large amounts of data \citep{li2018deeplearning}. The availability of large text corpora is especially relevant in the field of machine translation where the state-of-the-art approach to neural machine translation \citep{vaswani2017attention} requires large amounts of \emph{parallel texts}, i.e., texts in one language and their translation into another language. Parallel texts have also proven useful to build pre-trained language models with cross-lingual capabilities~\citep{conneau2020unsupervised,kale-etal-2021-nmt5,reid-artetxe-2022-paradise}, and in translation-memory tools \citep{bowker2002computer} to assist professional translators. The reduced availability of parallel documents, particularly for low-resource language pairs, is fuelling a growing interest in web mining, which has allowed to build some of the largest parallel corpora to date~\citep{el2020ccaligned,banon2020paracrawl,schwenk2021ccmatrix,banon2022macocu}.

State-of-the-art tools for harvesting parallel data from the Internet, like Bitextor~\citep{banon2020paracrawl,espla-gomis2016bitextors} and ILSP-FocusedCrawler~\citep{papavassiliou2018discovering}, use a web crawler to automatically browse the web and collect textual data. Web crawlers start with a list of seed URLs. The corresponding documents are downloaded and parsed, and any new URLs linked from them are added to a list of pending downloads. Crawling ends when the process is interrupted or when the list of pending downloads is exhausted. Finally, downloaded documents are processed to identify parallel content. This process is extremely inefficient, as many downloaded documents end up being discarded, either because they are not in a language of interest or because they are not parallel~\citep{banon2020paracrawl}. Furthermore, in large-scale crawls, where some websites are only partially crawled, this unguided approach leads to the omission of a portion of the available parallel content.
This situation is exacerbated for low-resourced language pairs, as the amount of parallel documents available on the Internet is scarcer.

Focused crawling aims at addressing this problem by prioritizing those documents that are more likely to be useful for the selected downstream task \citep{chakrabarti1999focused}. The literature includes several works that rely on the extracted content \citep{agarwal2014focused}, as well as works that use only the extracted URLs from downloaded documents to prioritize crawling \citep{baykan2013comprehensive,hernandez2016cala,han2018focused}. For the task of harvesting parallel data, we propose a smart bilingual focused crawler that ranks the URLs to be downloaded during crawling to prioritize the most promising documents, i.e. those that are likely to be parallel and in the desired languages. Our approach integrates the information provided by two models: one that infers the language in which a document is from its URL, and another that determines if two URLs point to two parallel documents, both without access to the documents' content. We build these models on top of XLM-RoBERTa~\citep{conneau2020unsupervised}, a multilingual model supporting  100 languages, which we fine-tuned on publicly available datasets.

We evaluate both standalone models and their integration into a crawling tool. The language identifier is evaluated on 161 languages, the parallelness identifier on 11 language pairs, and the integration of both models into a crawling tool on 4 low-resource language pairs. The results show that both models  perform their respective tasks with a reasonable accuracy, even though they only use the documents' URLs. In addition, their integration into a crawling tool leads to more parallel data with fewer documents crawled.

Our primary contribution to the state of the art is the introduction of the first focused-crawling approach designed explicitly for gathering parallel data. This approach enables the crawler to discover the parallel content in a website sooner, therefore reducing the bandwidth, and potentially the time, required for acquiring parallel corpora. Additionally, we introduce what, to the best of our knowledge, is the first approach in the literature to assess the likelihood of parallel content in documents based solely on their URLs without relying on \emph{ad~hoc} or naive rules~\citep{dara2016yoda,el2020ccaligned}. 

The rest of the paper is organized as follows. Next section reviews the related work in the literature. Sections~\ref{se:langid} and \ref{se:classifier} then describe the two models aforementioned and evaluate them in isolation. Section~\ref{se:smart-crawling} outlines the integration of these models into our smart bilingual focus crawling approach, which is subsequently evaluated in real crawling tasks. Finally, Section~\ref{se:concluding} ends with some concluding remarks.

\section{Related Work}\label{se:related}

As discussed above, conducting a general-purpose crawling for mining specific data is a waste of bandwidth and computing resources, as many downloaded documents end up being discarded for being irrelevant to the intended purpose. This challenge is addressed by developing focused crawlers \citep{chakrabarti1999focused} capable of identifying web pages relevant to the specific task at hand. This involves pruning URLs found in documents not relevant for the specific task and, optionally, ranking the URLs to prioritize downloading the most relevant ones first.
 
In the literature, focused crawlers are employed for various purposes, including hate detection \citep{agarwal2014focused}, medical sentiment analysis \citep{abbasi2013crawling}, and, most prominently, topic identification \citep{shrivastava2023efficient}. Numerous URL-based approaches have been proposed to guide crawling by identifying the topic of downloaded web pages: \citet{rajalakshmi2013web} use $n$-gram models to extract features from URLs and apply different supervised methods; \citet{hernandez2016cala} identify URL patterns given tuples of URLs and their associated topics; \citet{han2018focused} apply reinforcement learning to prioritize downloading URLs with a higher probability of leading to documents on given topics.

Gathering parallel content from the web necessitates document alignment. However, most approaches rely on content-based methods 
\citep{buck2016quick,guo2019hierarchical,thompson2020exploiting} and cannot be used to prioritize URLs before downloading documents. 
Contrary to content-based methods, URL-based approaches offer resource savings if the necessary information can be derived from URL elements. \citet{resnik2003web} and \citet{chen2000parallel} employ string substitution with language-dependent rules to guide crawling, while \citet{zhang2013finding} identify patterns from URL pairs differing in at least one character that may lead to parallel content. Hybrid approaches combining URL and content-based features to align parallel documents have also been proposed \citep{espla2010combining}.

Closer to our work, \citet{barbosa2011crawling} and \citet{baykan2013comprehensive} use URL-based machine-learning models for finding bilingual websites, and for language identification, respectively. \citet{barbosa2011crawling} detect patterns in individual URLs identifying websites that might lead to parallel documents, and process the content of the documents to identify the language ---unlike our approach which only uses the URLs--- and, if at least one document is found for each language of interest, they consider the  website  bilingual. \citet{baykan2013comprehensive} use $n$-grams from URLs to identify the language of documents and guide the crawling process; however they only support a small set of five languages, in contrast to the 161 languages we support.

\section{Language Identification from URLs}\label{se:langid}\label{se:langid-approach}

We approach the inference of the language of a document from its URL as a multi-class classification problem, and propose a model that utilizes XLM-RoBERTa \citep{conneau2020unsupervised} as a base to produce a probability distribution over the supported languages, including an \textit{unknown} class. 

\begin{figure}[t]
    \includegraphics[width=0.5\textwidth]{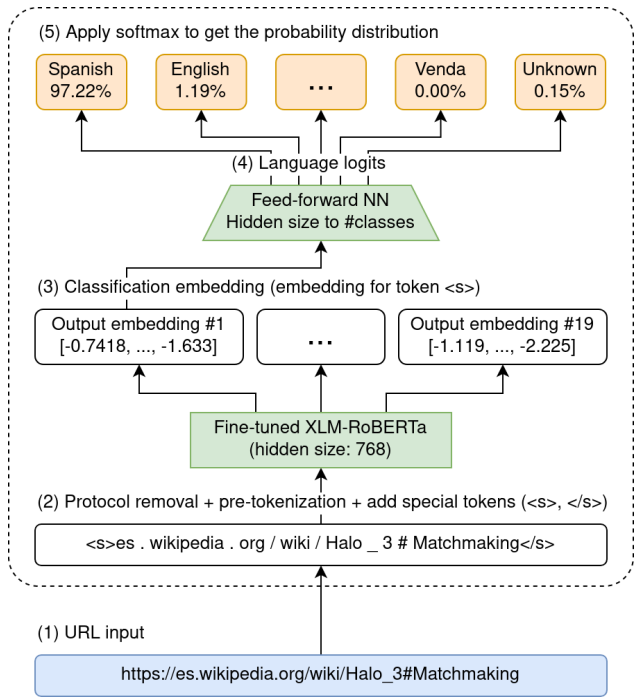}
    \caption{Architecture of the model used for language identification from URLs.}
    \label{fig:exp-langid-model-pipeline}
\end{figure}

Our approach works as follows. Firstly, the input URLs undergo pre-processing to remove the protocol, decode HTML entities, pre-tokenize by splitting groups of alphabetic characters, blanks, and special characters such as underscores, and to add special tokens \texttt{<s>} and \texttt{</s>} to delimit each URL. The URLs are then processed by the XLM-RoBERTa transformer (after applying its internal sub-word tokenizer) to generate the output embeddings. In a BERT fashion~\citep{devlin-etal-2019-bert}, the embedding of the first token (\texttt{<s>}) is used to represent each URL and passed to a feed-forward layer which is connected to a softmax output layer to obtain the final probability distribution over the set of supported languages. Figure~\ref{fig:exp-langid-model-pipeline} illustrates this architecture.\footnote{Code, models, and datasets are available at \url{https://github.com/transducens/url2lang/}}

\subsection{Experimental Settings}\label{se:langid-dataset}\label{se:langid-metrics}\label{se:langid-baseline}

\paragraph{Dataset.} 
We built a dataset consisting of pairs $(u,l)$ associating to each URL~$u$ the language~$l$ in which the linked document is written. 
We build on CommonCrawl\footnote{\url{https://commoncrawl.org/}} (CC) snapshots CC-MAIN-2023-06 and CC-MAIN-2023-14, which contain massive collections of web documents with language automatically identified using CLD2.\footnote{\url{https://github.com/CLD2Owners/cld2/}} While language is not manually annotated, CLD2 shows a very high performance with both macro precision and macro recall above $98\%$~\citep{cldeval14}. Independent studies confirm a significant macro F1 even off-the-shelf \citep{lui2014accurate}. 

The data in CC cover 161 languages that are represented following a Zipfian distribution.\footnote{\url{https://commoncrawl.github.io/cc-crawl-statistics/plots/languages}}  In order to deal with over-represented languages, a maximum of 600,000 URLs per language are collected, resulting in a corpus of about 50.4 million URLs. We split the data and use 50 million URLs for training, and the rest for development and testing (about 220,000 URLs each). We divide our data in such a way that URLs from the same web domain only appear in one of the splits. Figure~\ref{fig:langid-dataset-langs-size} reports the language distribution of each set.

\begin{figure}
\resizebox{\textwidth}{\height - 4cm - 16pt}{
\begin{tikzpicture}
\begin{axis}[
    xbar stacked,
    enlarge y limits=0.01,
    bar width=2pt,
    height=22.5cm,
    width=1.1*\linewidth,
    legend style={at={(0.5,-0.06)},
      anchor=north,legend columns=-1},
    xlabel={Data samples (log scale)},
    symbolic y coords={lif, kas, sux, ven, got, ssw, tso, nau, zha, chr, aka, nso, ipk, ton, run, fij, tsn, crs, aym, wol, bis, aar, mfe, dzo, sag, lug, iku, lin, kha, ile, syr, glv, vol, orm, abk, smo, sna, ina, sco, grn, sot, que, nya, haw, ibo, bih, hmn, kal, tir, yor, xho, mri, roh, sun, bod, uig, gla, zul, snd, yid, hat, kin, div, ceb, tuk, cos, war, bre, bak, mlg, fry, asm, pus, san, oci, hau, jav, amh, ltz, mlt, som, fao, lao, ori, tgk, kir, kur, pan, gle, tgl, afr, swa, sin, tat, glg, kaz, uzb, aze, guj, eus, cym, mon, mar, sqi, nno, mya, kat, ell, isl, swe, mkd, epo, slk, bul, lat, khm, bos, tha, hun, jpn, cat, hye, nor, heb, pol, ind, ron, zho, lav, lit, ara, nld, fas, spa, vie, est, ces, unk, eng, msa, rus, srp, deu, fra, hrv, tur, urd, ita, ukr, kor, fin, por, kan, dan, hin, mal, ben, tam, nep, slv, bel, tel},
    ytick=data,
    y tick label style={rotate=0,anchor=east,font=\tiny},
    ]

\addplot+[xbar, error bars/.cd, y dir=both, y explicit] plot coordinates {(3.044522437723423,lif) (4.442651256490317,kas) (5.3981627015177525,sux) (5.71042701737487,ven) (6.075346031088684,got) (6.09807428216624,ssw) (6.42648845745769,tso) (6.432940092739179,nau) (6.858565034791365,zha) (6.937314081223682,chr) (6.905753276311464,aka) (6.891625897052253,nso) (7.022868086082641,ipk) (7.085064293952548,ton) (7.138073034044347,run) (7.2485040723706105,fij) (7.237778191923443,tsn) (7.4205789054108005,crs) (7.433666540166168,aym) (7.582738488914411,wol) (7.622174594817622,bis) (7.687538766201629,aar) (7.795234929002173,mfe) (7.824046010856292,dzo) (7.880804344674901,sag) (8.10349427838097,lug) (8.149601735736155,iku) (8.200288260287554,lin) (8.254528881939745,kha) (8.465268118551318,ile) (8.555836815008442,syr) (8.594154232552366,glv) (9.063694791634697,vol) (9.072227069846548,orm) (9.086249986745132,abk) (9.127393451155257,smo) (9.177093778182547,sna) (9.1986725671085,ina) (9.213734605044198,sco) (9.241354425505353,grn) (9.266437111328052,sot) (9.314520190885846,que) (9.308283446306147,nya) (9.315240781801553,haw) (9.328656598794225,ibo) (9.3789010135201,bih) (9.40878131065002,hmn) (9.470702633773001,kal) (9.531118671917154,tir) (9.705524299752216,yor) (9.747885820441555,xho) (9.864226838582825,mri) (9.875705171523677,roh) (9.895455380838865,sun) (10.013731126634887,bod) (10.04207489229217,uig) (10.079874869841891,gla) (10.083974093384352,zul) (10.174277757317691,snd) (10.184258296685131,yid) (10.302498544237826,hat) (10.368227352106764,kin) (10.515316211918485,div) (10.531856183153893,ceb) (10.572393268723038,tuk) (10.578725312552505,cos) (10.571265641745532,war) (10.641536394447272,bre) (10.668141109747323,bak) (10.707505041927497,mlg) (10.723421557166176,fry) (10.718033446526634,asm) (10.813780332606648,pus) (10.823710542937889,san) (10.836419055558409,oci) (10.87492912887513,hau) (10.934712822502382,jav) (10.954641290044536,amh) (11.056461755890249,ltz) (11.088644442604423,mlt) (11.08217331763518,som) (11.10784028468035,fao) (11.140425970684525,lao) (11.280779008853019,ori) (11.290569103740589,tgk) (11.318332926695176,kir) (11.428510713827865,kur) (11.561639435760828,pan) (11.5742651252034,gle) (11.653139479058614,tgl) (11.679253132009833,afr) (11.725557968258775,swa) (11.745290368626037,sin) (11.756639975788797,tat) (11.876039632537863,glg) (11.914349083918673,kaz) (11.909914754597349,uzb) (11.912895502796413,aze) (11.921705074046223,guj) (11.91921023706207,eus) (11.91837057287839,cym) (11.915840658140155,mon) (11.922077102803021,mar) (11.91972301831217,sqi) (11.918190553075727,nno) (11.919410053231859,mya) (11.919350112573092,kat) (11.922993297604966,ell) (11.924266608650546,isl) (11.9221103129268,swe) (11.919975982319405,mkd) (11.918610548881942,epo) (11.919030368365732,slk) (11.924346136783665,bul) (11.925736856274575,lat) (11.919742991481549,khm) (11.92363015579287,bos) (11.920654674731459,tha) (11.923358213893902,hun) (11.922628248103518,jpn) (11.926001536079728,cat) (11.926160310342754,hye) (11.925796415338986,nor) (11.925511823346014,heb) (11.92506160551243,pol) (11.924809924880575,ind) (11.927442810997828,ron) (11.923974951372294,zho) (11.927522086944226,lav) (11.924843044372393,lit) (11.920741141659576,ara) (11.92839370781794,nld) (11.921279728779087,fas) (11.926239688021791,spa) (11.923809199105252,vie) (11.924140676170074,est) (11.923729628257316,ces) (11.92424672562899,unk) (11.924803300850582,eng) (11.922435713787001,msa) (11.92554491959985,rus) (11.924935773114395,srp) (11.923749521562888,deu) (11.924843044372393,fra) (11.92416718958918,hrv) (11.92351740837019,tur) (11.926199999969874,urd) (11.92892159338461,ita) (11.92653729821416,ukr) (11.92764759434107,kor) (11.930878934530753,fin) (11.926272760195198,por) (11.928631290808115,kan) (11.923875503308963,dan) (11.926266145848018,hin) (11.92952172464011,mal) (11.924359390857637,ben) (11.926874482726703,tam) (11.923835721314289,nep) (11.925650820251901,slv) (11.930417953291112,bel) (11.932516990413292,tel)}; 

\addplot+[xbar, error bars/.cd, y dir=both, y explicit] plot coordinates {(1.7918,lif) (2.0794,kas) (2.1972,sux) (1.0986,ven) (1.3863,got) (0.6931,ssw) (1.0986,tso) (0.6931,nau) (0.6931,zha) (1.0986,chr) (1.7918,aka) (3.9512,nso) (3.5553,ipk) (1.9459,ton) (2.9957,run) (1.3863,fij) (1.3863,tsn) (2.1972,crs) (2.1972,aym) (4.3820,wol) (1.3863,bis) (2.8332,aar) (4.5951,mfe) (0.6931,dzo) (1.6094,sag) (2.3026,lug) (4.6821,iku) (2.7726,lin) (3.4657,kha) (3.8712,ile) (3.1781,syr) (2.4849,glv) (2.9957,vol) (4.9345,orm) (0.6931,abk) (5.3033,smo) (6.0113,sna) (4.6151,ina) (4.8903,sco) (4.0604,grn) (5.1648,sot) (2.0794,que) (5.0752,nya) (4.1744,haw) (5.0173,ibo) (2.8332,bih) (7.1253,hmn) (5.5334,kal) (2.0794,tir) (6.5103,yor) (5.1475,xho) (5.9584,mri) (4.2627,roh) (5.3033,sun) (2.3979,bod) (6.8690,uig) (5.7652,gla) (5.2679,zul) (5.3083,snd) (5.2364,yid) (4.7005,hat) (6.8763,kin) (6.0259,div) (5.0106,ceb) (5.1874,tuk) (5.3471,cos) (6.0234,war) (2.8904,bre) (3.9512,bak) (5.6870,mlg) (4.8903,fry) (7.3185,asm) (6.2691,pus) (6.4800,san) (5.3033,oci) (5.5094,hau) (7.0211,jav) (6.1463,amh) (6.0753,ltz) (6.1696,mlt) (5.7652,som) (5.8579,fao) (4.9628,lao) (7.8713,ori) (7.6672,tgk) (4.2485,kir) (6.5958,kur) (6.2226,pan) (6.2538,gle) (7.1041,tgl) (5.9054,afr) (6.0113,swa) (5.9454,sin) (6.3404,tat) (10.2556,glg) (8.3745,kaz) (5.9162,uzb) (8.4562,aze) (6.4677,guj) (6.9603,eus) (6.1883,cym) (8.3859,mon) (7.0656,mar) (8.3490,sqi) (7.5251,nno) (8.0014,mya) (7.2786,kat) (7.8872,ell) (7.9610,isl) (8.1848,swe) (7.7940,mkd) (7.0211,epo) (8.2438,slk) (7.9223,bul) (8.0637,lat) (7.7973,khm) (8.2308,bos) (7.8951,tha) (8.3816,hun) (7.9338,jpn) (7.9645,cat) (7.2248,hye) (8.0255,nor) (7.6074,heb) (7.8316,pol) (8.0140,ind) (8.0346,ron) (7.8951,zho) (7.5066,lav) (8.0163,lit) (7.8954,ara) (7.9291,nld) (7.9099,fas) (7.9241,spa) (7.8540,vie) (8.0388,est) (7.7558,ces) (7.9273,unk) (7.8906,eng) (7.5979,msa) (7.9080,rus) (7.9161,srp) (7.9645,deu) (7.8797,fra) (7.8176,hrv) (7.7952,tur) (6.8046,urd) (7.8617,ita) (7.4518,ukr) (7.7021,kor) (7.6714,fin) (7.8095,por) (5.2730,kan) (7.6953,dan) (7.6596,hin) (6.7226,mal) (7.5766,ben) (6.7979,tam) (7.5914,nep) (7.4950,slv) (5.7170,bel) (6.6983,tel)}; 

\addplot+[xbar, error bars/.cd, y dir=both, y explicit] plot coordinates {(0.6931,lif) (2.3026,kas) (2.6391,sux) (4.2627,ven) (1.0986,got) (1.0986,ssw) (0.6931,tso) (1.3863,nau) (1.0986,zha) (1.0986,chr) (1.3863,aka) (0.6931,nso) (1.6094,ipk) (2.0794,ton) (4.2341,run) (2.9957,fij) (2.0794,tsn) (1.7918,crs) (1.6094,aym) (2.3026,wol) (5.2883,bis) (2.8904,aar) (2.0794,mfe) (3.6376,dzo) (1.7918,sag) (2.8904,lug) (2.0794,iku) (2.1972,lin) (2.8904,kha) (3.7612,ile) (2.3026,syr) (3.1781,glv) (2.7081,vol) (2.3979,orm) (1.3863,abk) (5.7683,smo) (5.7462,sna) (3.9318,ina) (4.8363,sco) (3.6889,grn) (5.8021,sot) (2.0794,que) (5.6168,nya) (5.7494,haw) (5.8636,ibo) (1.6094,bih) (5.2933,hmn) (5.2883,kal) (2.7081,tir) (6.0283,yor) (5.7557,xho) (5.7462,mri) (2.8332,roh) (6.0868,sun) (6.0162,bod) (4.7536,uig) (6.4135,gla) (6.4232,zul) (6.5309,snd) (5.8972,yid) (4.9836,hat) (6.0707,kin) (7.4668,div) (5.0499,ceb) (5.1874,tuk) (5.5872,cos) (5.2364,war) (3.5553,bre) (3.1781,bak) (6.0331,mlg) (5.7557,fry) (6.0186,asm) (6.2461,pus) (6.4846,san) (5.9054,oci) (6.0981,hau) (6.4036,jav) (7.1164,amh) (6.3404,ltz) (6.3682,mlt) (7.1452,som) (4.7875,fao) (6.7754,lao) (8.0298,ori) (5.5984,tgk) (6.8112,kir) (4.7875,kur) (7.7681,pan) (7.6695,gle) (7.3633,tgl) (7.8379,afr) (8.1730,swa) (7.5501,sin) (4.7005,tat) (7.4570,glg) (8.5019,kaz) (8.8445,uzb) (8.6408,aze) (6.4846,guj) (8.4160,eus) (7.5380,cym) (7.5606,mon) (7.7289,mar) (8.2723,sqi) (6.0403,nno) (7.9313,mya) (7.8713,kat) (8.5775,ell) (7.8043,isl) (8.0883,swe) (8.1997,mkd) (7.2869,epo) (8.1233,slk) (8.3335,bul) (8.0665,lat) (7.9530,khm) (7.5994,bos) (8.0060,tha) (7.7350,hun) (8.2191,jpn) (7.7350,cat) (7.6866,hye) (7.9638,nor) (7.9255,heb) (8.2177,pol) (7.8969,ind) (7.9662,ron) (8.0944,zho) (7.8071,lav) (7.6852,lit) (7.9821,ara) (8.0258,nld) (7.9728,fas) (8.0014,spa) (7.9673,vie) (7.5590,est) (8.0974,ces) (7.9463,unk) (7.9714,eng) (7.9113,msa) (7.9291,rus) (7.7824,srp) (7.8450,deu) (7.9183,fra) (7.8047,hrv) (7.9610,tur) (7.6774,urd) (7.8973,ita) (8.0414,ukr) (7.9402,kor) (7.8586,fin) (7.6871,por) (7.7289,kan) (7.8857,dan) (7.6266,hin) (7.9277,mal) (7.5491,ben) (7.7660,tam) (6.9354,nep) (7.3072,slv) (6.6796,bel) (6.0684,tel)}; 

\legend{\strut Training, \strut Development, \strut Testing}
\end{axis}
\end{tikzpicture}
}
\caption{Language distribution for the training, development and test sets used in the experiments for language indentification.}
    \label{fig:langid-dataset-langs-size}
\end{figure}

\paragraph{Evaluation Metric.}
We present precision, recall, and F1 results, along with their aggregated counterparts: macro precision, macro recall, and macro F1. These aggregated metrics represent the average values across all possible classes and offer a comprehensive overview of the systems' performance.

\paragraph{Baseline Systems.}
We compare our approach to a baseline that, inspired by \citet{el2020ccaligned} and \citet{baykan2013comprehensive}, looks at different components of the URL that may indicate the language of the linked document: subdomain, public suffix, directory names, and value of the different parameters (if any). The value of each component is compared to the ISO 639-1 and ISO 639-2 language codes.

To address potential contradictions from various URL components, we examine them in a specific order established through evaluation on the development set. The optimal sequence, determined by assessing all possible combinations, is as follows: parameter values, directory names, public suffix, and subdomain.

Additionally, we include a baseline simpler than our transformer-based model (see Section~\ref{se:langid}) using FastText~\citep{joulin-etal-2017-bag}, a common approach for language identification \citep{fasttext-langid,nllb}. This system is a linear classifier that represents text as a bag of character-level $n$-grams. The embeddings of these $n$-grams (extracted from each word independently) are averaged to form a hidden representation, which is then fed into a hierarchical softmax classifier to compute the probability distribution over the supported classes.

For this baseline, hereafter referred to as FastText, we use the same training data and data pre-processing pipeline as our approach. This ensures that both models support the same label set ---enabling a direct comparison--- and effectively adapts the model to classify URLs rather than text. Regarding hyperparameters, we adopt those used by the publicly available language identification model \textit{lid.176.bin}~\citep{fasttext-langid}; namely, 16-dimensional embeddings, character $n$-grams from $2$ to $4$, learning rate of $0.1$, and $10$ epochs.

\subsection{Training of the Model}\label{se:langid-training}
We used HuggingFace Transformers \citep{wolf-etal-2020-transformers} and PyTorch \citep{paszke2019pytorch} to fine-tune XLM-RoBERTa$_\mathrm{Base}$ using the cross-entropy loss function. Based on preliminary experiments, we set the learning rate to $10^{-5}$ and used the AdamW optimizer \citep{loshchilov2017decoupled} with default parameters ($\beta_1=0.9$, $\beta_2=0.999$, $\epsilon=10^{-8}$, $\lambda=0.01$). 
We use a learning rate scheduler with an initial warm-up phase ($10\%$ of the training steps in the first epoch) where the learning rate linearly increases to the configured value and then decays following an inverse square root pattern.

Due to the extensive data in CC, utilizing the entire training set in each training epoch is impractical. Therefore, we opted to incrementally process the training set in folds. Starting with one million URLs, we add an additional million URLs after each training epoch. Training stops when further data fails to enhance the macro F1 metric on the development set for five epochs. The training data shown in Figure~\ref{fig:langid-dataset-langs-size} correspond to the data actually used for training.

Finally, following  \citet{zhang2020revisiting} ---who argue that layers near the output of a pretrained model retain task-specific knowledge--- 
we experimented with re-initializing the top $n\in[0,5]$ layers of the transformer. Our findings reveal that reinitializing the top layer ($n=1$) before fine-tuning for our downstream task slightly improves the results on the development set. Therefore, we adopted this configuration.

\subsection{Results and Discussion}\label{se:langid-evaluation}

\begin{figure*}
\centering
\makebox[\textwidth][c]{
\begin{subfigure}[t]{0.5\textwidth}
\centering
\begin{tikzpicture}
\begin{axis}[
	xlabel=F1,
    legend style={at={(0.5,-0.06)},
    anchor=north,legend columns=-1},
    enlarge y limits=0.01,
    ytick=data,
    height=19.0cm,
    xmin=0,
    xmax=101,
    xbar=0.4pt,
    bar width=1.4pt,
    ytick distance=1pt,
    width=\linewidth,
    grid=none,
    y tick label style={rotate=0,font=\tiny},
    symbolic y coords={ibo,uzb,yor,zul,hmn,yid,mri,haw,mlg,sna,nya,isl,smo,lit,gla,xho,hat,sot,hun,fin,fao,tgk,ceb,ron,slk,gle,snd,amh,pol,mlt,mkd,lav,eus,nor,nld,pus,fry,tat,tgl,bul,hau,ita,guj,sun,hrv,tel,glg,bel,cat,kir,deu,rus,tuk,aze,fra,afr,ltz,urd,spa,cos,epo,kin,lao,swa,ukr,por,est,sin,som,ind,mon,kat,tha,ara,mal,kan,mar,khm,cym,sqi,mya,kaz,pan,hin,hye,slv,msa,heb,jav,swe,nep,ell,ben,ces,tam,kor,dan,eng,srp,fas,tur,bos,ori,vie,jpn,nno,zho,lat,oci,san,unk,sco,war},
    legend to name={smart-crawling-experiment-langid-per-lang-legend},
    legend entries={Baseline,FastText,Ours},
]
\addplot+[fill=blue!30, draw=blue, xbar, error bars/.cd, y dir=both, y explicit]
	plot coordinates {(99.86,ibo) (99.54,uzb) (99.39,yor) (99.27,zul) (98.98,hmn) (98.89,yid) (98.37,mri) (98.21,haw) (98.05,mlg) (98.04,sna) (97.97,nya) (97.41,isl) (97.22,smo) (96.97,lit) (96.69,gla) (96.21,xho) (94.93,hat) (94.78,sot) (94.53,hun) (93.30,fin) (93.27,fao) (92.97,tgk) (92.73,ceb) (92.61,ron) (92.42,slk) (92.37,gle) (91.13,snd) (90.00,amh) (88.03,pol) (86.63,mlt) (85.12,mkd) (84.68,lav) (84.42,eus) (82.98,nor) (82.94,nld) (79.95,pus) (79.62,fry) (78.26,tat) (77.60,tgl) (76.67,bul) (76.39,hau) (74.64,ita) (73.43,guj) (69.83,sun) (67.69,hrv) (66.36,tel) (66.14,glg) (65.16,bel) (62.26,cat) (57.95,kir) (56.55,deu) (55.50,rus) (54.96,tuk) (54.69,aze) (53.83,fra) (52.56,afr) (50.86,ltz) (45.45,urd) (44.83,spa) (43.27,cos) (42.84,epo) (41.32,kin) (40.76,lao) (34.79,swa) (33.79,ukr) (32.80,por) (32.11,est) (27.93,sin) (27.49,som) (24.69,ind) (22.54,mon) (22.50,kat) (22.16,tha) (22.12,ara) (22.10,mal) (21.52,kan) (21.16,mar) (18.85,khm) (18.53,cym) (17.64,sqi) (17.57,mya) (17.19,kaz) (15.08,pan) (14.71,hin) (14.47,hye) (14.21,slv) (13.84,msa) (12.59,heb) (12.15,jav) (11.67,swe) (10.15,nep) (9.16,ell) (8.66,ben) (8.02,ces) (7.98,tam) (7.58,kor) (6.88,dan) (6.69,eng) (6.60,srp) (6.02,fas) (5.68,tur) (5.35,bos) (4.51,ori) (4.40,vie) (2.70,jpn) (1.89,nno) (1.15,zho) (0.75,lat) (0.54,oci) (0.00,san) (0.00,unk) (0.00,sco) (0.00,war)};
\addlegendimage{fill=red!30, draw=red, xbar legend}
\addplot+[fill=brown!30, draw=brown, xbar, error bars/.cd, y dir=both, y explicit]
	plot coordinates {(99.86,ibo) (99.52,uzb) (99.15,yor) (99.11,zul) (98.98,hmn) (98.62,yid) (98.72,mri) (97.58,haw) (98.06,mlg) (98.54,sna) (97.62,nya) (98.42,isl) (98.44,smo) (98.29,lit) (97.69,gla) (95.61,xho) (94.04,hat) (96.66,sot) (97.49,hun) (96.39,fin) (93.62,fao) (98.72,tgk) (97.37,ceb) (97.35,ron) (96.60,slk) (98.93,gle) (99.35,snd) (73.35,amh) (96.66,pol) (96.85,mlt) (94.64,mkd) (96.11,lav) (98.79,eus) (91.27,nor) (91.65,nld) (97.46,pus) (81.23,fry) (68.18,tat) (91.51,tgl) (86.66,bul) (78.51,hau) (90.80,ita) (88.18,guj) (77.35,sun) (80.54,hrv) (80.81,tel) (89.05,glg) (98.17,bel) (92.08,cat) (99.61,kir) (79.76,deu) (82.23,rus) (93.99,tuk) (91.10,aze) (85.65,fra) (98.10,afr) (85.09,ltz) (71.19,urd) (83.29,spa) (80.21,cos) (97.73,epo) (96.51,kin) (99.66,lao) (97.70,swa) (91.62,ukr) (90.72,por) (95.76,est) (89.71,sin) (98.22,som) (78.29,ind) (66.64,mon) (99.41,kat) (57.90,tha) (68.76,ara) (86.90,mal) (91.43,kan) (89.39,mar) (74.03,khm) (98.35,cym) (91.76,sqi) (36.72,mya) (98.69,kaz) (31.96,pan) (44.16,hin) (95.07,hye) (91.58,slv) (85.99,msa) (87.54,heb) (83.78,jav) (94.98,swe) (62.56,nep) (97.64,ell) (84.57,ben) (94.55,ces) (79.73,tam) (83.70,kor) (95.17,dan) (44.98,eng) (77.41,srp) (91.11,fas) (76.12,tur) (77.91,bos) (7.21,ori) (96.29,vie) (77.73,jpn) (47.97,nno) (83.77,zho) (61.23,lat) (91.70,oci) (77.77,san) (70.97,unk) (34.17,sco) (33.33,war)};
\end{axis}
\end{tikzpicture}
\end{subfigure}
~ \hspace{0pt}
\begin{subfigure}[t]{0.5\textwidth}
\centering
\begin{tikzpicture}
\begin{axis}[
	xlabel=F1,
    legend style={at={(0.5,-0.06)},
    anchor=north,legend columns=-1},
    enlarge y limits=0.01,
    ytick=data,
    height=19.0cm,
    xmin=0,
    xmax=101,
    xbar=0.4pt,
    bar width=1.2pt,
    ytick distance=1pt,
    width=\linewidth,
    grid=none,
    y tick label style={rotate=0,font=\tiny},
    symbolic y coords={ibo,uzb,yor,zul,hmn,yid,mri,haw,mlg,sna,nya,isl,smo,lit,gla,xho,hat,sot,hun,fin,fao,tgk,ceb,ron,slk,gle,snd,amh,pol,mlt,mkd,lav,eus,nor,nld,pus,fry,tat,tgl,bul,hau,ita,guj,sun,hrv,tel,glg,bel,cat,kir,deu,rus,tuk,aze,fra,afr,ltz,urd,spa,cos,epo,kin,lao,swa,ukr,por,est,sin,som,ind,mon,kat,tha,ara,mal,kan,mar,khm,cym,sqi,mya,kaz,pan,hin,hye,slv,msa,heb,jav,swe,nep,ell,ben,ces,tam,kor,dan,eng,srp,fas,tur,bos,ori,vie,jpn,nno,zho,lat,oci,san,unk,sco,war},
    legend to name={smart-crawling-experiment-langid-per-lang-legend},
    legend entries={Baseline,FastText,Ours},
]
\addlegendimage{fill=blue!30, draw=blue, xbar legend}
\addplot+[fill=red!30, draw=red, xbar, error bars/.cd, y dir=both, y explicit]
    plot coordinates {(60.69,afr) (35.25,amh) (68.57,ara) (92.88,aze) (78.97,bel) (74.56,ben) (64.31,bos) (84.6,bul) (80.31,cat) (88.16,ceb) (91.85,ces) (49.15,cos) (97.22,cym) (88.79,dan) (65.58,deu) (94.24,ell) (30.05,eng) (81.97,epo) (91.12,est) (94.64,eus) (89.47,fao) (87.15,fas) (94.64,fin) (76.8,fra) (76.14,fry) (93.1,gla) (86.31,gle) (79.73,glg) (73.96,guj) (78.37,hat) (60.38,hau) (90.77,haw) (57.22,heb) (58.13,hin) (98.23,hmn) (75.9,hrv) (92.67,hun) (76.31,hye) (95.84,ibo) (63.27,ind) (95.47,isl) (78.15,ita) (49.96,jav) (61.92,jpn) (44.53,kan) (95.68,kat) (95.59,kaz) (70.4,khm) (81.69,kin) (96.38,kir) (70.51,kor) (63.17,lao) (35.69,lat) (93.18,lav) (97.27,lit) (80.31,ltz) (78.27,mal) (77.06,mar) (92.14,mkd) (81.47,mlg) (92.33,mlt) (61.83,mon) (77.44,mri) (69.92,msa) (23.74,mya) (60.64,nep) (75.64,nld) (35.56,nno) (82.48,nor) (86.21,nya) (76.71,oci) (7.84,ori) (31.69,pan) (91.44,pol) (80.81,por) (83.42,pus) (94.82,ron) (71.3,rus) (80.29,san) (11.49,sco) (79.78,sin) (93.27,slk) (73.23,slv) (84.52,smo) (84.94,sna) (93.22,snd) (94.4,som) (72.85,sot) (72.61,spa) (81.35,sqi) (73.01,srp) (38.06,sun) (73.76,swa) (89.8,swe) (53.77,tam) (60.87,tat) (59.46,tel) (94.66,tgk) (87.78,tgl) (67.45,tha) (79.0,tuk) (84.78,tur) (88.52,ukr) (49.73,unk) (77.95,urd) (98.88,uzb) (91.78,vie) (14.88,war) (92.38,xho) (95.16,yid) (92.87,yor) (60.8,zho) (94.43,zul)};
\addplot+[fill=brown!30, draw=brown, xbar, error bars/.cd, y dir=both, y explicit]
	plot coordinates {(99.86,ibo) (99.52,uzb) (99.15,yor) (99.11,zul) (98.98,hmn) (98.62,yid) (98.72,mri) (97.58,haw) (98.06,mlg) (98.54,sna) (97.62,nya) (98.42,isl) (98.44,smo) (98.29,lit) (97.69,gla) (95.61,xho) (94.04,hat) (96.66,sot) (97.49,hun) (96.39,fin) (93.62,fao) (98.72,tgk) (97.37,ceb) (97.35,ron) (96.60,slk) (98.93,gle) (99.35,snd) (73.35,amh) (96.66,pol) (96.85,mlt) (94.64,mkd) (96.11,lav) (98.79,eus) (91.27,nor) (91.65,nld) (97.46,pus) (81.23,fry) (68.18,tat) (91.51,tgl) (86.66,bul) (78.51,hau) (90.80,ita) (88.18,guj) (77.35,sun) (80.54,hrv) (80.81,tel) (89.05,glg) (98.17,bel) (92.08,cat) (99.61,kir) (79.76,deu) (82.23,rus) (93.99,tuk) (91.10,aze) (85.65,fra) (98.10,afr) (85.09,ltz) (71.19,urd) (83.29,spa) (80.21,cos) (97.73,epo) (96.51,kin) (99.66,lao) (97.70,swa) (91.62,ukr) (90.72,por) (95.76,est) (89.71,sin) (98.22,som) (78.29,ind) (66.64,mon) (99.41,kat) (57.90,tha) (68.76,ara) (86.90,mal) (91.43,kan) (89.39,mar) (74.03,khm) (98.35,cym) (91.76,sqi) (36.72,mya) (98.69,kaz) (31.96,pan) (44.16,hin) (95.07,hye) (91.58,slv) (85.99,msa) (87.54,heb) (83.78,jav) (94.98,swe) (62.56,nep) (97.64,ell) (84.57,ben) (94.55,ces) (79.73,tam) (83.70,kor) (95.17,dan) (44.98,eng) (77.41,srp) (91.11,fas) (76.12,tur) (77.91,bos) (7.21,ori) (96.29,vie) (77.73,jpn) (47.97,nno) (83.77,zho) (61.23,lat) (91.70,oci) (77.77,san) (70.97,unk) (34.17,sco) (33.33,war)};
\end{axis}
\end{tikzpicture}
\end{subfigure}
}

\ref*{smart-crawling-experiment-langid-per-lang-legend}
\caption{Language identification results on a per-language basis, comparing our model with the baseline (left subfigure) and with a FastText model trained on the same data (right subfigure). Only languages with a minimum of 100 URLs and 10 different web domains in the test set are included.}
    \label{fig:exp-langid-eval-isolation-all-langs}
\end{figure*}

Table~\ref{table:exp-langid-eval-isolation} shows the results of the baseline system, the FastText system, and our model on the test set: our model surpasses the baseline and FastText systems across all metrics. While the baseline attains a notable macro precision score, its macro recall is low. This reduced recall is attributed to the absence of language information in some URLs, even in multilingual websites. This is corroborated by the significant percentage of test samples classified as \textit{unknown} by the baseline (57.4\%). The FastText system achieves better results than the baseline in terms of macro F1 and exhibits more balanced performance across the reported metrics, similar to our transformer-based model; however, it still falls behind our approach by more than 10 points in macro precision, macro recall, and macro F1.

\begin{table}[tb]
\begin{tabular}{ lrrr}
\toprule
 \headrow System & Macro Precision & Macro Recall & Macro F1 \\
\midrule
Baseline & 65.29\% & 34.42\% & 38.24\% \\
FastText & 60.92\% & 57.32\% & 57.41\% \\
 Ours & \textbf{72.59\%} & \textbf{69.04\%} & \textbf{69.22\%} \\
\bottomrule
\end{tabular}
\caption{Results of the baseline, FastText, and our language identifier. 
Refer to Figure~\ref{fig:exp-langid-eval-isolation-all-langs} for a breakdown per language.}
\label{table:exp-langid-eval-isolation}
\end{table}

Figure~\ref{fig:exp-langid-eval-isolation-all-langs} offers a comprehensive breakdown of the F1 metric per language, revealing the consistent superiority of our model over the baseline (left subfigure) across all languages, except for a few instances with comparable F1 ---notably the baseline achieves better results in Amharic and Tatar. The FastText system (right subfigure) yields results closer to our approach across languages; however, the overall tendency indicates that our model consistently outperforms it, with the only exceptions being Sanskrit, Oriya, Turkish, Hindi, Thai, Urdu, and Azerbaijani.

It is noteworthy that all the systems ---baseline, FastText, and our model--- yield unexpectedly low results for English; the best performing model (ours) achieves a precision of $36.6\%$ and a recall of $58.5\%$. Further analysis reveals a pattern of frequent confusion between non-English URLs and English (28 languages are confused with English in at least $10\%$ of their data). These findings align with those reported by the authors of CLD2, confirming a bias towards English, especially for Interlingua (CLD2 classifies $44.4\%$ of these documents as English). Noticeably, our model behaves similarly and $46\%$ of the URLs marked as Interlingua were misclassified as English.

Another source of noise in the data stems from URLs linking non-English documents, but that incorporate English text in the URL, such as the Spanish webpage \textit{https://www.dochaney.com/es/contact-us}. This issue is exacerbated when the URL lacks any explicit language hint, as seen in the Spanish webpage 
\textit{https://hjmorelia.com/about-us}. 
Upon manual inspection of the URLs in the test set linking to English documents, we found that $94.1\%$ of these URLs lack any language identifier mark (e.g., ``/en/'' or ``/*-eng/''), compelling our model to rely on detecting the presence of English words or other intricate patterns, if any. This fact, along with the presence of webpages that link to non-English documents but contain English text in their URLs, such as the aforementioned examples of URLs linking to Spanish webpages, may be an additional reason why the model struggles with URLs linking to English documents.

The problems encountered are inherently related to the use of URLs to infer information about the content of the linked document before it is actually downloaded. URLs are often useful because they present similar semantics to the documents they link to. However, identifying the language of documents from their URLs is not always feasible due to insufficient or inaccurate semantics. During our analysis, we encountered several examples of URLs that appeared to link to a document in a particular language but instead led to a different one, as illustrated above.

\section{Inferring Parallelness from URLs}\label{se:classifier}

We model the task of determining whether two URLs point to parallel documents as a binary classification problem. Our model takes a pair of URLs as input and outputs the probability of the linked documents being parallel.

\begin{figure}
    \includegraphics[width=0.5\textwidth]{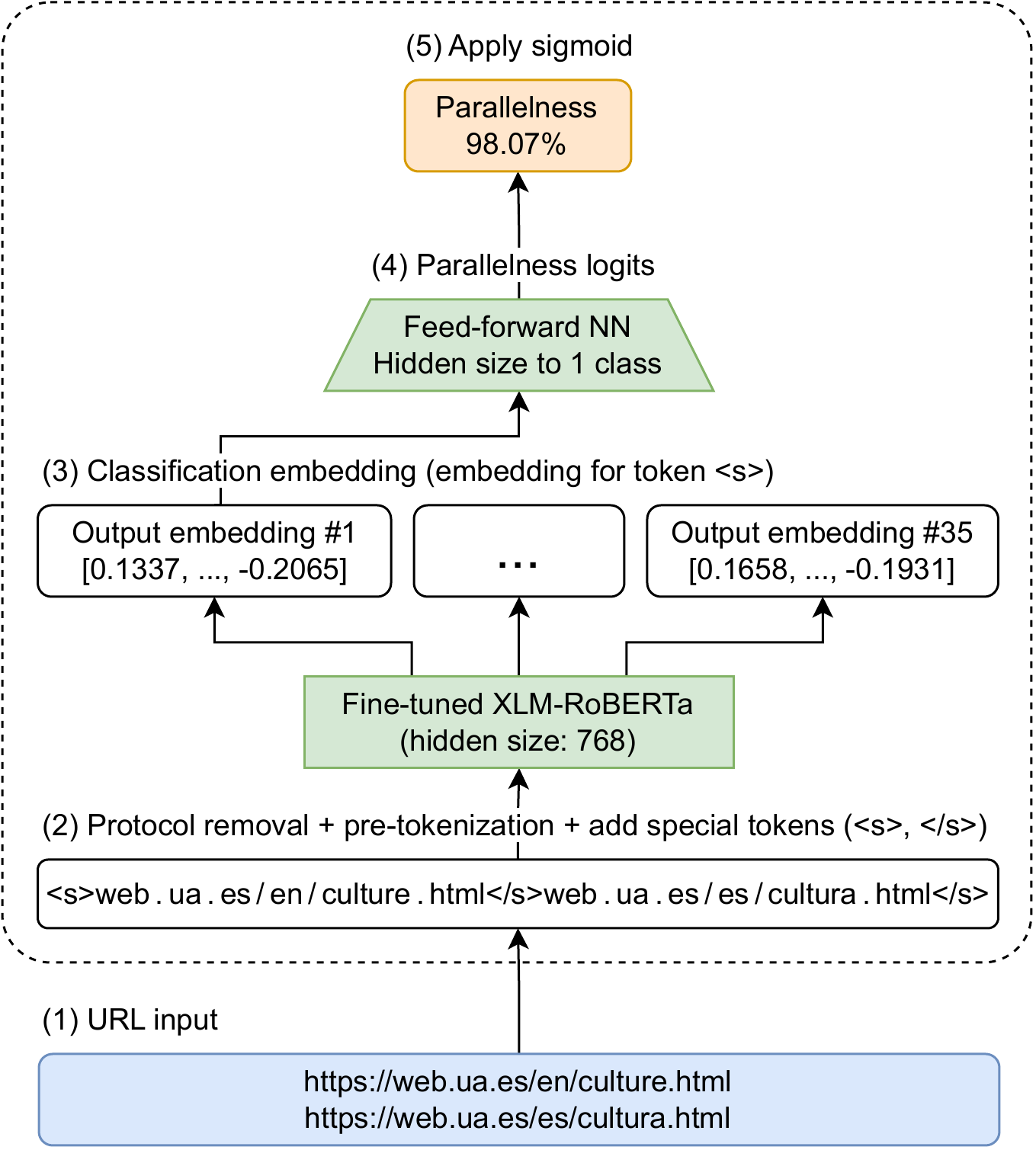}
    \caption{Architecture of the model used for inferring parallelness from URL pairs.}
    \label{fig:exp-classifier-model-pipeline}
\end{figure}

The approach mirrors that outlined in Section~\ref{se:langid-approach}. Differences lie in the architecture: the model takes two URLs as input, and the output embedding for the first token is projected through a feed-forward layer to a single output neuron whose output is normalized using a sigmoid function, representing the probability of the two input URLs pointing to parallel documents. Both URLs undergo identical pre-processing and are separated by the special token \texttt{</s>}. Figure~\ref{fig:exp-classifier-model-pipeline} illustrates this architecture.\footnote{Code, models, and datasets are available at \url{https://github.com/transducens/parallel-urls-classifier/}}

\subsection{Experimental Settings}\label{se:classifier-dataset}

\paragraph{Dataset.}
Our model is trained on a labelled dataset of URL pairs, indicating parallel or non-parallel relationships. The most relevant publicly available dataset for this task is the one distributed for the WMT16 bilingual document alignment shared task~\citep{buck2016findings} (hereafter WMT16); it consists of a collection of  documents in English and French obtained from 252 multilingual websites. It is divided into training (documents from 49 websites) and test (documents from 203 websites) sets. For every document, plain text and HTML content is available, along with its URL and automatically-detected language. A gold standard consisting of 1,624 URL pairs for training, and 2,402 for testing, is also distributed. This dataset has three significant limitations: (i) it supports only one language pair; (ii) the gold standard is relatively small and includes only a subset of the actual parallel documents in the training and test sets; and (iii) the gold standard lacks negative samples, i.e., URL pairs labelled as non-parallel.\footnote{Notice that the WMT16 shared task was initially designed as a mining task, not a classification task.}

We deal with this last limitation by extending the original WMT16 gold standard with negative samples obtained by emulating the behaviour of a web crawler ---which is the scenario in which we plan to use our model--- as follows. For each URL in the gold standard we extract all the URLs linked in the corresponding HTML content and discard those pointing to documents neither in English, nor in French. We then build a set of URL pairs by pairing the original URL with every linked URL. Finally, we check if one of these pairs appears in the gold standard, and if it does, we label the rest of pairs as non-parallel; otherwise we discard all the pairs in the set. 

We alleviate the other two limitations by extending our training data with the  MaCoCu~\citep{banon2022macocu} v2 corpus, which covers 10 additional language pairs; namely, English aligned with Albanian, Bulgarian, Croatian, Icelandic, Macedonian, Maltese, Montenegrin, Slovenian, Serbian and Turkish.\footnote{We choose MaCocu v2 in front of other multilingual corpora crawled from the Internet such as ParaCrawl~\citep{banon2020paracrawl}, or CCAligned~\citep{el2020ccaligned} as it has been reported to be of higher quality \citep{macocu-v2-evaluation-techreport}.} This corpus is distributed aligned both at the segment level and at the document level with URLs attached to each pair of aligned documents; although these alignments are not as accurate as those in WMT16 because documents were automatically aligned. Additionally, this corpus does not include the original HTML content, which prevents the application of the method described above for generating negative samples. To address this limitation, we devise strategies inspired by those poposed by \citet{espla-gomis2016bitextors} and \citet{dara2016yoda} to generate negative samples without accessing the HTML content.

\paragraph{Creation of Synthetic Negative Samples.}\label{se:classifier-synthetic-negative-samples}
In what follows we evaluate various methods for creating synthetic negative samples from the URL pairs $(u_A,v_B)$ in the MaCoCu v2 corpus, where $u_A$ and $v_B$ point to documents in languages $A$ and $B$, respectively. These methods are:
\begin{description}
\item[Random match:] A new URL pair $(u_A, v_B')$ is generated by replacing $v_B$ with $v_B'$, a randomly selected URL from the pair collection. This approach is rather naive, as the resulting URLs are likely to be significantly different.
\item[Remove random tokens:] Both $u_A$ and $v_B$ are tokenized, and random tokens are removed; random tokens never include the scheme, authority, or the port.
\item[Maximize Jaccard similarity:] Similarly to \textit{random match}, a new URL pair $(u_A,v_B')$ is created, but in this case $v_B'$ is chosen so that it maximizes the Jaccard similarity to $v_B$. This method aims to create URL pairs very similar to the original one.
\end{description}
These methods assume that all URLs link documents in languages $A$ or $B$, which is the case for the WMT16 dataset. However, in a realistic crawling scenario, the language cannot be identified until the document is downloaded, and it is common to encounter pairs of URLs linking documents in the same language. Hence, it is crucial to include negative samples corresponding to non-parallel documents in the same language for training. We achieve this by generating synthetic monolingual negative samples where $A=B$, using pairs of identical URLs, $(u_A, u_A)$ and $(v_B, v_B)$, as starting points.

We evaluated all combinations of the strategies outlined above on the WMT16 dataset, for which genuine negative samples can be obtained from HTML, as discussed earlier; the comprehensive results obtained in this evaluation are available in~\ref{ap:synthetic}.
The results indicate that the most effective combination comprises \textit{random match} (both monolingual and bilingual), \textit{maximize Jaccard similarity} (both monolingual and bilingual), and \textit{remove random token} (bilingual only). To create a cohesive dataset for our experiments, we apply these strategies to generate synthetic negative samples from  both WMT16 and MaCoCu v2. 

We use $60\%$ of the web domains in the WMT16 training split for training, and the other $40\%$ for development.\footnote{Notice that there is no development set in the WMT16 dataset.} The web domains in MaCoCu v2 are split as follows: $80\%$ for training, $10\%$ for development, and $10\%$ for testing; we made sure that URLs from the same web domain only appear in one of these splits. Table~\ref{table:classifier-dataset-langs-size} reports the amount of positive and synthetic negative samples in each set per language pair.

\begin{table}[tb]
\begin{tabular}{lrrr}
\toprule
\headrow Pair & Train & Dev & Test \\
\midrule
eng-fra & 0.9k/5k  & 0.7k/5k  & 2k/15k   \\ 
eng-cnr & 29k/394k & 2k/20k   & 3k/34k   \\ 
eng-mlt & 29k/405k & 0.4k/6k  & 9k/125k  \\ 
eng-isl & 30k/400k & 4k/51k   & 9k/125k  \\ 
eng-mkd & 38k/518k & 6k/84k   & 23k/315k \\ 
eng-sqi & 54k/742k & 8k/113k  & 9k/127k  \\ 
eng-slv & 209k/3M  & 18k/245k & 31k/427k \\ 
eng-bul & 231k/3M  & 32k/437k & 38k/517k \\ 
eng-srp & 197k/3M  & 57k/729k & 51k/646k \\ 
eng-hrv & 246k/3M  & 49k/669k & 22k/292k \\ 
eng-tur & 444k/6M  & 66k/879k & 50k/669k \\ 
\bottomrule
\end{tabular}
\caption{Positive/synthetic negative samples per language pair in the training, development and test sets used to train the model for inferring the parallelness of two documents from their URLs.}
\label{table:classifier-dataset-langs-size}
\end{table}

\paragraph{Evaluation Metrics.}\label{se:classifier-metrics}
We assess our model on both the official WMT16 test set and the test set described in the previous paragraph. For the WMT16 test set, we employ recall and \emph{soft recall}~\citep{buck2016findings}, a variant of recall used in the shared task that considers near matches. Precision cannot be used due to the non-exhaustive nature of the WMT16 gold standard, rendering the actual correct URL pairs unavailable for the test set.

As regards the evaluation on the test set described above, the same metrics described in Section~\ref{se:langid-metrics} are used: precision, recall, F1, macro precision, macro recall, and macro F1.

\paragraph{Baseline System.}\label{se:classifier-baseline}
We use the official baseline of the WMT16 shared task\footnote{\url{https://github.com/christianbuck/wmt16-document-alignment-task/}}~\citep{buck2016findings}.\footnote{
None of the submissions to the WMT16 shared task can be considered as a baseline because all of them rely on the content of the documents to align them.} This baseline defines a set of tokens considered to be potential language identifiers in a URL; a given pair of URLs is aligned if they can be transformed into the same string by removing one or more of these tokens. We have extended this baseline to support all the languages covered in our experiments. Similar to the already supported languages, the set of tokens for additional languages includes the name of the language in English, its endonym, the ISO 639-1 and ISO 639-2 language codes, and combinations of the ISO 639-1 language codes and the ISO 3166-1 country codes where the language is official, separated by a hyphen; e.g. for Albanian: \textit{albanian}, \textit{shqip}, \textit{sqp}, \textit{alb}, \textit{sq}, \textit{sq-sq}, \textit{sq-ks}.

\subsection{Training of the Model}\label{se:classifier-training}
We employed the same training configuration as regards learning rate, optimizer, hyperparameters, and loss function, as described in Section~\ref{se:langid-training}. 
Training stops if the macro F1 computed on the development set does not improve after five epochs. In addition, the top layer of the XLM-RoBERTa$_\mathrm{Base}$ model used is re-initialized before training.

\subsection{Results and Discussion}\label{se:classifier-evaluation}
Table~\ref{table:exp-classifier-eval-isolation-wmt16} reports the results in terms of recall and soft recall on the WMT16 test set. The figures in this table can be directly compared to those achieved by the systems submitted to the shared task (\citet[Tables~3 \& 4]{buck2016findings}; 19 submissions from 11 distinct research groups). To compute them, we obtain the Cartesian product of all English and French URLs in the WMT16 test set and apply our model and the baseline to each URL pair. As WMT16 only allows 1-to-1 URL alignments, our model outputs the most probable alignment among those identified.
The results obtained are promising: our approach not only outperforms the baseline in almost 7 points of soft recall, but also four of the nineteen  systems submitted to the shared task. This is especially relevant given that all the systems submitted compare the content of documents to determine whether they are parallel or not, while our model solely relies on URLs.

\begin{table}[t]
\begin{tabular}{ lrr }
\toprule
 \headrow System & Recall & Soft Recall \\
\midrule
 Baseline & 59.91 & 59.91 \\
 Ours & \textbf{62.99} & \textbf{66.69} \\ 
 \bottomrule
\end{tabular}
\caption{Recall and soft recall obtained by our  parallel URL identifier and the baseline on the WMT16 test set.}
\label{table:exp-classifier-eval-isolation-wmt16}
\end{table}

\begin{table}[tb]
 \begin{tabular}{llrr}
 \toprule
 \headrow Class & Metric & Baseline & Ours \\
\midrule
  & Precision & 64.37 & \textbf{76.28} \\ 
  Positive & Recall & 12.78 & \textbf{61.79} \\ 
  & F1 & 21.33 & \textbf{68.28} \\ 
\midrule
  & Precision & 93.86 & \textbf{97.19} \\ 
  Negative & Recall & \textbf{99.47} & 98.57 \\ 
  & F1 & 96.58 & \textbf{97.87} \\ 
\midrule
  & Precision & 79.12 & \textbf{86.74} \\ 
  Macro & Recall & 56.13 & \textbf{80.18} \\ 
  & F1 & 58.96 & \textbf{83.08} \\ 
\bottomrule
\end{tabular}
\caption{Results for the parallelness identifier from URLs on the dataset described in Section~\ref{se:classifier-dataset}, considering all language pairs collectively.}
\label{table:exp-classifier-eval-isolation}
\end{table}

\begin{figure}[tb]
\begin{tikzpicture}
\begin{axis}[
	ylabel=Macro F1,
    legend style={at={(0.46,-0.14)}, anchor=north,legend columns=-1},
    enlarge x limits=0.06,
    ymin=0,
    ymax=105,
    ytick align=outside,
    ybar=0.5pt,
    bar width=7pt,
    xtick distance=1pt,
    width=0.7\linewidth,
    height=0.55\linewidth,
    grid=none,
    x tick label style={rotate=90,font=\footnotesize},
    symbolic x coords={fra,hrv,bul,cnr,mlt,isl,tur,slv,mkd,srp,sqi},
]
\addplot+[error bars/.cd, y dir=both, y explicit]
    coordinates {(bul,64.64) (cnr,62.15) (fra,86.09) (hrv,67.93) (isl,59.84) (mkd,52.93) (mlt,61.59) (slv,56.64) (sqi,50.54) (srp,51.88) (tur,59.57)};
\addplot+[error bars/.cd, y dir=both, y explicit]
	coordinates {(bul,80.39) (cnr,85.09) (fra,84.94) (hrv,87.74) (isl,81.51) (mkd,74.55) (mlt,79.88) (slv,83.42) (sqi,73.37) (srp,89.83) (tur,81.48)};
\legend{Baseline,Ours}
\end{axis}
\end{tikzpicture}
\caption{Macro F1 scores per language (paired with English) for the  parallelness identifier from URLs on the dataset described in Section~\ref{se:classifier-dataset}.}
    \label{fig:exp-classifier-eval-isolation-per-lang}
\end{figure}
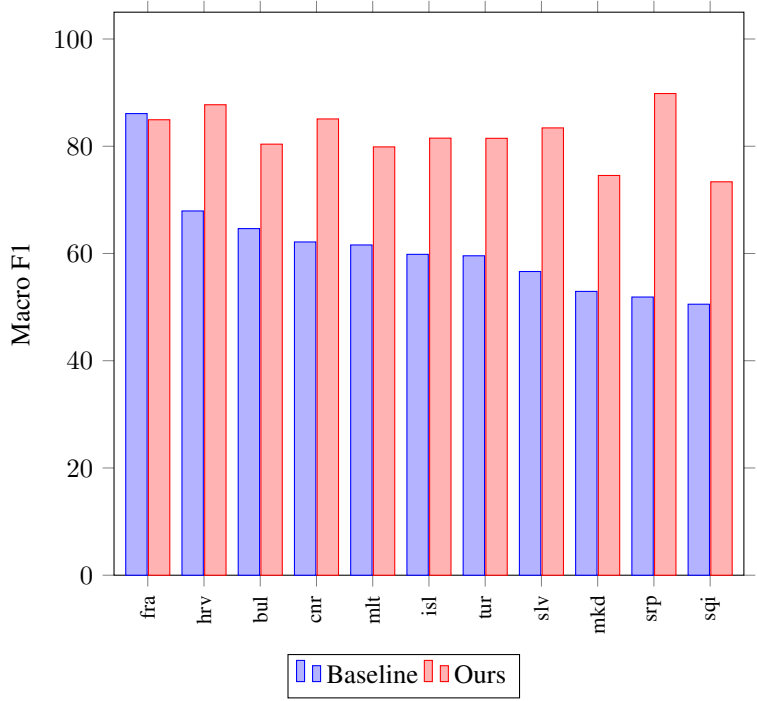

Table~\ref{table:exp-classifier-eval-isolation} presents the results on the test set described in Section~\ref{se:classifier-dataset} for all language pairs combined; Figure~\ref{fig:exp-classifier-eval-isolation-per-lang} displays the macro F1 score for each language pair individually. 
Table~\ref{table:exp-classifier-eval-isolation} reveals the baseline's poor performance for the positive class, with a low recall below $15\%$, in contrast to the high recall (above $99\%$) for the negative class. This suggests that the baseline's simplicity struggles with the complexities of parallel URLs. However, it achieves a competitive macro precision, although not surpassing our system. This is because it looks for explicit language hints in the URLs and if a URL pair is deemed parallel, it is likely to be a true positive, except for potential confusion with common strings in ISO-639 codes (e.g., ``isl'' confused with ``island'' instead of Icelandic).

Figure~\ref{fig:exp-classifier-eval-isolation-per-lang} shows that our model systematically outperforms, in terms of macro F1, the baseline in all languages, but French. Manual inspection confirmed a significant difference between the English--French test set, derived from the WMT16 dataset, and the rest of the test sets, derived from the MaCoCu v2 corpus. About 60\% of the positive samples in the English--French dataset exclusively use language identification tokens in URLs as the method to identify translated documents.\footnote{For example, pairs of URLs \texttt{www.un.org/en/} and \texttt{www.un.org/fr/}, that only differ in the tokens \textit{en} and \textit{fr}.} Conversely, for the datasets used for the rest of languages this percentage ranges from 10\% to 15\%. The fact that the baseline designed for the WMT16 task exclusively builds on language identification tokens to detect parallel documents explains the high performance on this specific test set. While the details about the methodology followed to build the WMT16 gold standard are not public, it seems likely that the procedure followed somehow biased this dataset towards over-representing websites using language identification tokens.

\section{Smart Bilingual Focused Crawling}\label{se:smart-crawling}

In this section we describe and evaluate our smart bilingual focus crawling strategy leveraging the models discussed in sections~\ref{se:langid} and \ref{se:classifier}. We implement our crawling approach on Heritrix 3,\footnote{\url{https://github.com/internetarchive/heritrix3/}} an open-source and widely used crawler.\footnote{We created a fork of Heritrix 3 to implement the required changes for our experiment. The code is available at \url{https://github.com/cgr71ii/heritrix3/tree/puc_uri_cost/}} Like other crawling tools, Heritrix operates with a list of pending downloads initially populated with a set of seed URLs. During the crawl, URLs are extracted from this list, and the downloaded documents are parsed to append any newly discovered URLs to it. By default, Heritrix explores the URLs in the list of pending downloads in the order in which they were added (i.e., breadth-first search). However, we modify this behavior by using a priority queue where the priority of each URL is determined by combining the probabilities obtained from the two models described above. Additionally, we ensure that the seed URLs are always prioritized for download in the first place.

\begin{figure*}[t]
\includegraphics[width=\textwidth]{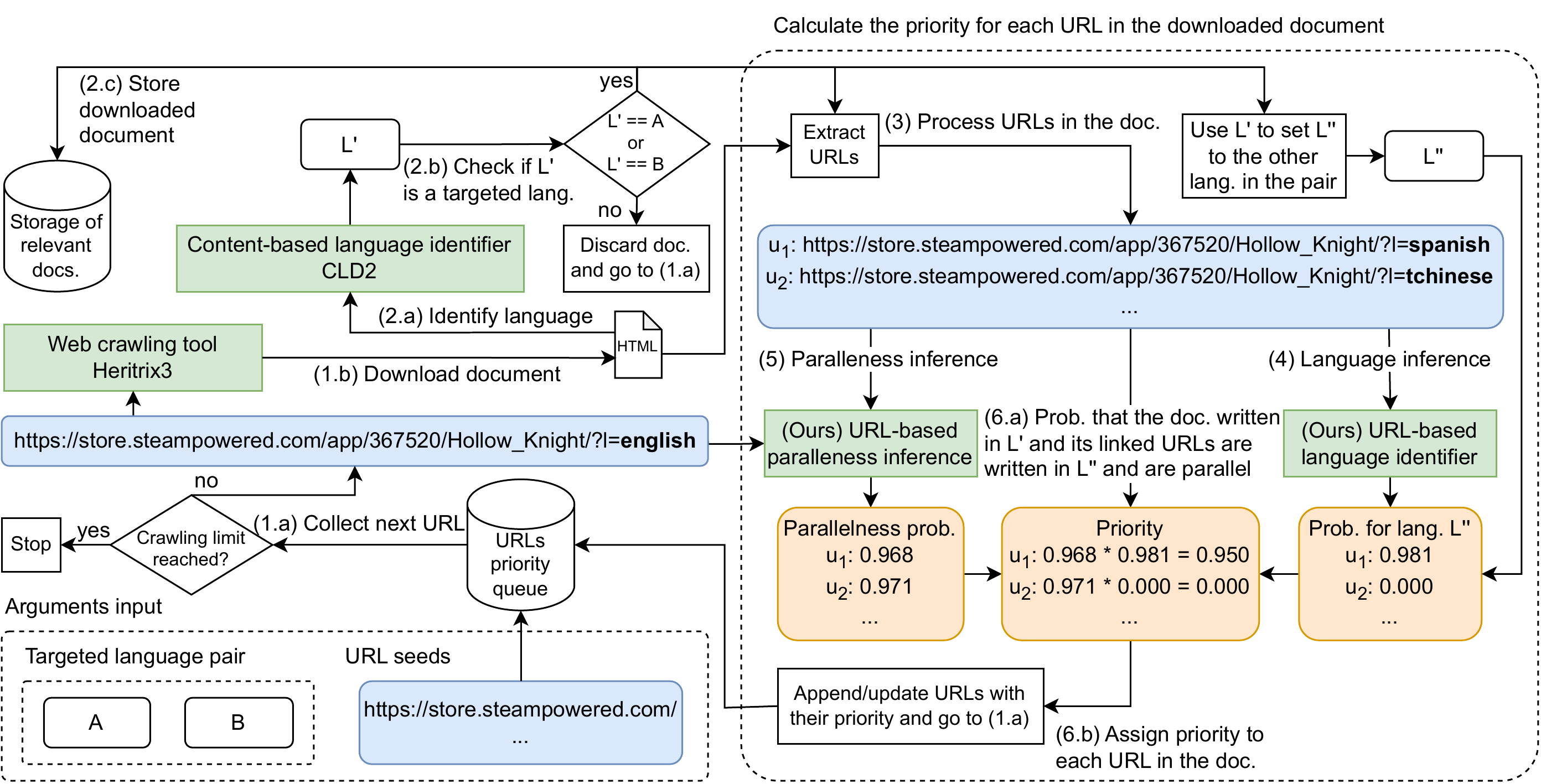}
\caption{General architecture of our approach for smart bilingual focused crawling of parallel documents. In this example, $A$ and $B$ are English and Spanish, the language $L'$ of the downloaded document is English, and the language $L''$ used to obtain the language probabilities is Spanish.}
\label{fig:general-architecture}
\end{figure*}

Figure~\ref{fig:general-architecture} illustrates the general architecture of our crawling approach. For a specified pair of languages $A$ and $B$ we proceed as follows:

\begin{enumerate}
    \item A URL $u$ is taken from the list of pending downloads and the corresponding document $D_u$ is downloaded.
    \item CLD2 is used to determine the language of $D_u$; if it is not either $A$ or $B$, the document is discarded, otherwise the document is stored and the process continues.
    \item $D_u$ is parsed and the collection of URLs $\{v_i\}_1^N$ linked from it is extracted.
    \item The model that identifies the language of a document from its URL is used to determine the probability of each URL $v_i$ being in language $A$ if $D_u$ is in language $B$, or in language $B$ if $D_u$ is in language $A$.
    \item The model that infers parallelness from URLs is used to obtain the probability of each pair $\{(u,v_i)\}_1^N$ of being parallel.
    \item Each URL $v_i$ is added to the list of pending downloads with a priority score determined by multiplying the probabilities obtained in steps 4 and 5. If $v_i$ is already in the list, its priority is updated only if the new priority is higher.
\end{enumerate}
Note that this process is based on the assumption that web pages containing the same content but in different languages are typically mutually linked.

\subsection{Experimental Settings}\label{se:smart-crawling-experiments}
\paragraph{Dataset.} We evaluate the crawling strategy described above on four language pairs: English--Icelandic (eng-isl), English--Maltese (eng-mlt), English--Finnish (eng-fin), and Spanish--Basque (spa-eus). It is worth noting that neither Finnish, Spanish, nor Basque were used to fine-tune the model for inferring parallelness from URLs, and Maltese was not even used to train the base model, XLM-RoBERTa, on which our models build. 

We built a collection of seed URLs for each of the four bilingual crawling tasks. To ensure that the websites to be crawled actually contained parallel data, we used the Paracrawl v9 parallel corpus to identify productive websites for the languages covered in our experiment. Paracrawl consists of parallel sentences for 43 language pairs, along with their corresponding source URLs. For each language pair, we built the list of unique URLs\footnote{Since some of the URLs in Paracrawl do not exist anymore, we discarded those with an HTTP connection status other than \textit{OK} (HTTP code 200).} and  grouped them by  websites; we then ranked websites by the amount of URLs in this list. Websites present in the training and development sets used to fine-tune the models described in sections~\ref{se:langid} and \ref{se:classifier} were excluded. Ultimately, a lists of seed URLs was created for each language pair by collecting the home page of the 200-top websites.\footnote{The home page of a website is defined as its URL without any additional resource (e.g., \textit{https://en.wikipedia.org} for \textit{https://en.wikipedia.org/wiki/Deep\_learning}).}

\paragraph{Evaluation.} We measure the performance of our approach in terms of the amount of parallel data downloaded at different moments of a crawling process. We include three models in our comparison: (a) the original implementation of Heritrix; (b) the proposed approach that integrates the two models proposed in Section~\ref{se:langid} and Section~\ref{se:classifier}, along with CLD2 (Heritrix+CLD2+smart); and (c) a crawler integrating only CLD2 to exclude documents not in the targeted languages (Heritrix+CLD2). The inclusion of the last crawler helps determine the individual contributions of the proposed models and CLD2. We run each version of Heritrix on each website included in the experiment for a maximum of 48 hours. We then evaluate the amount of parallel data obtained by deciles of the total data downloaded per website.

To measure the amount of parallel data acquired, we used Bitextor\footnote{\url{https://github.com/bitextor/bitextor/}} because manual assessment would have been prohibitively expensive. Although Bitextor may introduce some noise, this equally affects all evaluated approaches. In any case, it is important to note that Bitextor is a state-of-the-art tool for harvesting parallel content \citep{banon2020paracrawl,espla-gomis2016bitextors} widely used and that has been utilized to create popular datasets such as ParaCrawl \citep{banon2020paracrawl} and MaCoCu \citep{banon2022macocu}.

\subsection{Results and Discussion}\label{se:smart-crawling-evaluation}

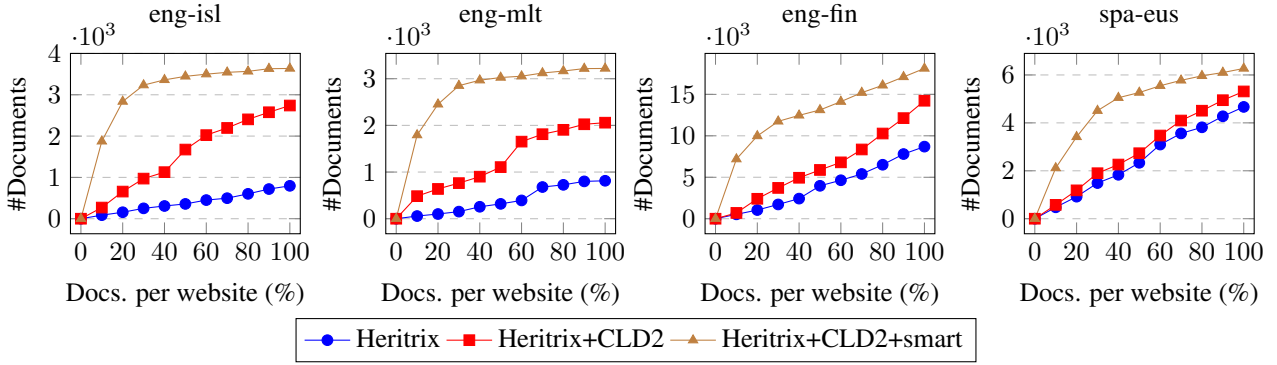
\begin{figure*}
    \makebox[\textwidth][c]{
    \begin{subfigure}[t]{0.31\textwidth}
        \centering
        \begin{tikzpicture}
            \begin{axis}[
                title={eng-isl},
                xlabel={\small Docs. per website (\%)},
                ylabel={\small \#Documents},
                ylabel style={yshift=-17pt},
                xmin=-5, xmax=105,
                xtick={0,20,40,60,80,100},
                ymajorgrids=true,
                grid style=dashed,
                width=\textwidth,
                scaled y ticks=base 10:-3,
                legend to name={smart-crawling-experiment-results-per-lang-legend},
                legend columns=-1,
                legend entries={Heritrix,Heritrix+CLD2,Heritrix+CLD2+smart},
            ]
        
            \addplot[
                color=blue,
                mark=*,
                ]
                coordinates {
                (0,0)(10,86)(20,157)(30,251)(40,308)(50,358)(60,451)(70,497)(80,602)(90,718)(100,796)
                };
        
            \addplot[
                color=red,
                mark=square*,
                ]
                coordinates {
                (0,0)(10,270)(20,657)(30,971)(40,1127)(50,1674)(60,2023)(70,2197)(80,2405)(90,2577)(100,2740)
                };

            \addplot[
                color=brown,
                mark=triangle*,
                ]
                coordinates {
                (0,0)(10,1874)(20,2837)(30,3234)(40,3361)(50,3446)(60,3499)(70,3543)(80,3567)(90,3631)(100,3637)
                };
        
            \end{axis}
        \end{tikzpicture}
    \end{subfigure}
    ~ \hspace{-28pt}
    \begin{subfigure}[t]{0.31\textwidth}
        \centering
        \begin{tikzpicture}
            \begin{axis}[
                title={eng-mlt},
                xlabel={\small Docs. per website (\%)},
                ylabel style={yshift=-15pt},
                xmin=-5, xmax=105,
                xtick={0,20,40,60,80,100},
                ymajorgrids=true,
                grid style=dashed,
                width=\textwidth,
                scaled y ticks=base 10:-3,
            ]
        
            \addplot[
                color=blue,
                mark=*,
                ]
                coordinates {
                (0,0)(10,57)(20,101)(30,152)(40,256)(50,318)(60,392)(70,679)(80,726)(90,800)(100,813)
                };
        
            \addplot[
                color=red,
                mark=square*,
                ]
                coordinates {
                (0,0)(10,483)(20,638)(30,763)(40,901)(50,1106)(60,1652)(70,1811)(80,1905)(90,2024)(100,2058)
                };

            \addplot[
                color=brown,
                mark=triangle*,
                ]
                coordinates {
                (0,0)(10,1793)(20,2452)(30,2851)(40,2968)(50,3026)(60,3054)(70,3122)(80,3165)(90,3217)(100,3220)
                };
        
            \end{axis}
        \end{tikzpicture}
    \end{subfigure}
    ~ \hspace{-37pt}
    \begin{subfigure}[t]{0.31\textwidth}
        \centering
        \begin{tikzpicture}
            \begin{axis}[
                title={eng-fin},
                xlabel={\small Docs. per website (\%)},
                ylabel style={yshift=-12pt},
                xmin=-5, xmax=105,
                xtick={0,20,40,60,80,100},
                ytick={0,5000,10000,15000},
                ymajorgrids=true,
                grid style=dashed,
                scaled y ticks=base 10:-3,
                width=\textwidth,
                legend to name={smart-crawling-experiment-results-per-lang-legend},
                legend columns=-1,
                legend entries={Heritrix,Heritrix+CLD2,Heritrix+CLD2+smart},
            ]
        
            \addplot[
                color=blue,
                mark=*,
                ]
                coordinates {
                (0,0)(10,527)(20,1047)(30,1720)(40,2425)(50,3969)(60,4640)(70,5397)(80,6510)(90,7800)(100,8695)
                };
        
            \addplot[
                color=red,
                mark=square*,
                ]
                coordinates {
                (0,0)(10,698)(20,2404)(30,3716)(40,4940)(50,5882)(60,6799)(70,8353)(80,10277)(90,12133)(100,14230)
                };

            \addplot[
                color=brown,
                mark=triangle*,
                ]
                coordinates {
                (0,0)(10,7190)(20,9978)(30,11746)(40,12468)(50,13117)(60,14124)(70,15206)(80,16091)(90,17103)(100,18124)
                };
        
            \end{axis}
        \end{tikzpicture}
    \end{subfigure}
    ~ \hspace{-33pt}
    \begin{subfigure}[t]{0.31\textwidth}
        \centering
        \begin{tikzpicture}
            \begin{axis}[
                title={spa-eus},
                xlabel={\small Docs. per website (\%)},
                ylabel style={yshift=-15pt},
                xmin=-5, xmax=105,
                xtick={0,20,40,60,80,100},
                ymajorgrids=true,
                grid style=dashed,
                width=\textwidth,
                scaled y ticks=base 10:-3,
            ]
        
            \addplot[
                color=blue,
                mark=*,
                ]
                coordinates {
                (0,0)(10,476)(20,920)(30,1488)(40,1833)(50,2334)(60,3095)(70,3560)(80,3815)(90,4275)(100,4666)
                };
        
            \addplot[
                color=red,
                mark=square*,
                ]
                coordinates {
                (0,0)(10,578)(20,1184)(30,1899)(40,2261)(50,2733)(60,3472)(70,4098)(80,4508)(90,4946)(100,5312)
                };

            \addplot[
                color=brown,
                mark=triangle*,
                ]
                coordinates {
                (0,0)(10,2116)(20,3415)(30,4509)(40,5049)(50,5266)(60,5549)(70,5775)(80,5963)(90,6100)(100,6271)
                };
        
            \end{axis}
        \end{tikzpicture}
    \end{subfigure}
    }

    \ref*{smart-crawling-experiment-results-per-lang-legend}
    \caption{
    For the three crawlers (Heritrix, Heritrix+CLD2 and Heritrix+CLD2+smart) and the language pairs eng-isl, eng-mlt, eng-fin, and spa-eus, thousands of parallel documents retrieved (y-axis) as a function of the percentage of documents downloaded (x-axis) from each website. Each data point accumulates the number of parallel documents downloaded from each website up to the percentage indicated on the x-axis.
    }
    \label{fig:smart-crawling-experiment-results-per-lang}
\end{figure*}

Figure~\ref{fig:smart-crawling-experiment-results-per-lang} illustrates the results obtained by the three models; the x-axis corresponds to the percentage\footnote{The reason for using relative rather than absolute values on the x-axis is that the number of documents downloaded from each website may vary due to external factors such as server workload or network speed.} of the total number of documents downloaded for each website after 48 hours of crawling, while the y-axis represents the amount of parallel data identified at each moment. The number of parallel documents for a given percentage of downloaded data was computed separately for each website, and the figure shows the aggregated results. It is worth noting that the number of parallel documents identified can serve as a proxy for crawling speed, avoiding the drawbacks of measuring crawling times, such as server workload or computing resource constraints. If more parallel documents are identified with the same amount of downloaded documents, then we are potentially improving crawling speed and bandwidth usage with respect to the downloaded documents.

The results obtained for all language pairs confirm that the Heritrix+CLD2+smart approach is capable of identifying most of the available parallel data earlier than the other two approaches. As expected, the difference between the three approaches diminishes as a higher fraction of the available data is downloaded, and they would eventually converge if the crawling time was enough to allow crawling all the data in these websites. Another notable observation is that the trend is consistent across all language pairs, regardless of whether they were seen during training, partially confirming the zero-shot capabilities of our approach (Heritrix+CLD2+smart). Notice that while our approach for inferring parallelness is language-independent, the model used to identify the language of documents from their URLs (see Section~\ref{se:langid}) was fine-tuned on all languages included in the experiments. Hence, a more extensive evaluation on a larger number of language pairs is required to robustly confirm the zero-shot capabilities of our method.

\section{Concluding Remarks}\label{se:concluding}

In order to optimize the crawling of parallel documents from the Internet and save bandwidth
we have introduced two models that work solely on the URLs and that are then integrated into a crawling tool. The first model is able to infer the language in which a document is from its URL;
the second one determines if two URLs point to two parallel documents.  We have thoroughly evaluated the two models in isolation, and they have demonstrated utility for their respective tasks. 

We have integrated these two models into a crawling tool to prioritize downloading URLs that are more likely to lead to parallel content in the desired languages. Experiments with four different language pairs demonstrate a clear increase in the ratio of parallel documents to downloaded documents, thereby reducing the number of discarded documents. This improves bandwidth usage and has the potential to reduce crawling time while yielding a greater quantity of parallel documents compared to regular crawling approaches. Additionally, the approach proves effective with language pairs not seen by the model used for inferring parallelness from URLs.

We have not assessed multilingual language models other than XLM-RoBERTa. Exploring newer models such as mT5 \citep{xue-etal-2021-mt5} could provide further evidence regarding the effectiveness of our approach but would require greater computational resources due to the trend of increasing the number of model parameters \citep{smith2022using}. Nevertheless, our main conclusion remains unaffected: URL-based models built on top of a multilingual language model can effectively guide crawling to obtain more parallel documents than traditional approaches.

All code, datasets derived from the corpora used, and models used in our experiments are made publicly available to facilitate replication and validation by other researchers, as well as for their use for smart bilingual focus crawling.

\paragraph{Acknowledgments.}

This work was supported by the Spanish Ministry of Science and Innovation (MCIN), the Spanish Research Agency (AEI/10.13039/501100011033) and the European Regional Development Fund A way to make Europe [grant number PID2021-127999NB-I00]; Cristian García-Romero is funded by Generalitat Valenciana and the European Social Fund [research grant CIACIF/2021/365]. The computational resources used were funded by the European Regional Development Fund through project IDIFEDER/2020/003.

\paragraph{Competing Interests.}

The author(s) declare none.

\printbibliography

@Book{li2018deeplearning,
author = {Deng, Li and Liu, Yang},
title = {Deep Learning in Natural Language Processing},
year = {2018},
isbn = {9789811052088},
publisher = {Springer Singapore},
address = {Singapore},
edition = {1st}
}

@inproceedings{banon2020paracrawl,
    title = "{P}ara{C}rawl: Web-Scale Acquisition of Parallel Corpora",
    author = "Ba{\~n}{\'o}n, Marta  and
      Chen, Pinzhen  and
      Haddow, Barry  and
      Heafield, Kenneth  and
      Hoang, Hieu  and
      Espl{\`a}-Gomis, Miquel  and
      Forcada, Mikel L.  and
      Kamran, Amir  and
      Kirefu, Faheem  and
      Koehn, Philipp  and
      Ortiz Rojas, Sergio  and
      Pla Sempere, Leopoldo  and
      Ram{\'\i}rez-S{\'a}nchez, Gema  and
      Sarr{\'\i}as, Elsa  and
      Strelec, Marek  and
      Thompson, Brian  and
      Waites, William  and
      Wiggins, Dion  and
      Zaragoza, Jaume",
    booktitle = "Proceedings of the 58th Annual Meeting of the Association for Computational Linguistics",
    month = jul,
    year = "2020",
    address = "Online",
    publisher = "Association for Computational Linguistics",
    url = "https://aclanthology.org/2020.acl-main.417",
    doi = "10.18653/v1/2020.acl-main.417",
    pages = "4555--4567",
}

@inproceedings{banon2022macocu,
    title = "{M}a{C}o{C}u: Massive collection and curation of monolingual and bilingual data: focus on under-resourced languages",
    author = "Ba{\~n}{\'o}n, Marta  and
      Espl{\`a}-Gomis, Miquel  and
      Forcada, Mikel L.  and
      Garc{\'\i}a-Romero, Cristian  and
      Kuzman, Taja  and
      Ljube{\v{s}}i{\'c}, Nikola  and
      van Noord, Rik  and
      Sempere, Leopoldo Pla  and
      Ram{\'\i}rez-S{\'a}nchez, Gema  and
      Rupnik, Peter  and
      Suchomel, V{\'\i}t  and
      Toral, Antonio  and
      van der Werff, Tobias  and
      Zaragoza, Jaume",
    booktitle = "Proceedings of the 23rd Annual Conference of the European Association for Machine Translation",
    month = jun,
    year = "2022",
    address = "Ghent, Belgium",
    publisher = "European Association for Machine Translation",
    url = "https://aclanthology.org/2022.eamt-1.41",
    pages = "303--304",
}

@Book{bowker2002computer,
  author =       {Bowker, Lynne},
  title =        {Computer-aided translation technology: a practical introduction},
  publisher =    {University of Ottawa Press},
  year =         2002
}

@InProceedings{vaswani2017attention,
  title={Attention is all you need},
  author={Vaswani, Ashish and Shazeer, Noam and Parmar, Niki and Uszkoreit, Jakob and Jones, Llion and Gomez, Aidan N and Kaiser, {\L}ukasz and Polosukhin, Illia},
  booktitle={Proceedings of the 31st International Conference on Neural Information Processing Systems},
  address={Long Beach, USA},
  pages={6000--6010},
  year={2017}
}

@inproceedings{espla-gomis2016bitextors,
    title = "Bitextor{'}s participation in {WMT}{'}16: shared task on document alignment",
    author = "Espl{\`a}-Gomis, Miquel  and
      Forcada, Mikel  and
      Ortiz-Rojas, Sergio  and
      Ferr{\'a}ndez-Tordera, Jorge",
    booktitle = "Proceedings of the First Conference on Machine Translation: Volume 2, Shared Task Papers",
    month = aug,
    year = "2016",
    address = "Berlin, Germany",
    publisher = "Association for Computational Linguistics",
    url = "https://aclanthology.org/W16-2367",
    doi = "10.18653/v1/W16-2367",
    pages = "685--691",
}

@inproceedings{papavassiliou2018discovering,
    title = "Discovering Parallel Language Resources for Training {MT} Engines",
    author = "Papavassiliou, Vassilis  and
      Prokopidis, Prokopis  and
      Piperidis, Stelios",
    booktitle = "Proceedings of the Eleventh International Conference on Language Resources and Evaluation ({LREC} 2018)",
    month = may,
    year = "2018",
    address = "Miyazaki, Japan",
    publisher = "European Language Resources Association (ELRA)",
    url = "https://aclanthology.org/L18-1599",
}

@inproceedings{conneau2020unsupervised,
    title = "Unsupervised Cross-lingual Representation Learning at Scale",
    author = "Conneau, Alexis  and
      Khandelwal, Kartikay  and
      Goyal, Naman  and
      Chaudhary, Vishrav  and
      Wenzek, Guillaume  and
      Guzm{\'a}n, Francisco  and
      Grave, Edouard  and
      Ott, Myle  and
      Zettlemoyer, Luke  and
      Stoyanov, Veselin",
    booktitle = "Proceedings of the 58th Annual Meeting of the Association for Computational Linguistics",
    month = jul,
    year = "2020",
    address = "Online",
    publisher = "Association for Computational Linguistics",
    url = "https://aclanthology.org/2020.acl-main.747",
    doi = "10.18653/v1/2020.acl-main.747",
    pages = "8440--8451",
}

@article{baykan2013comprehensive,
  title={A comprehensive study of techniques for URL-based web page language classification},
  author={Baykan, Eda and Henzinger, Monika and Weber, Ingmar},
  journal={ACM Transactions on the Web (TWEB)},
  volume={7},
  number={1},
  pages={1--37},
  year={2013},
  publisher = {Association for Computing Machinery},
  address = {New York, NY, USA},
  url = {https://dl.acm.org/doi/pdf/10.1145/2435215.2435218}
}

@article{chakrabarti1999focused,
  title = {Focused crawling: a new approach to topic-specific Web resource discovery},
  journal = {Computer Networks},
  volume = {31},
  number = {11},
  pages = {1623--1640},
  year = {1999},
  issn = {1389-1286},
  doi = {https://doi.org/10.1016/S1389-1286(99)00052-3},
  url = {https://www.sciencedirect.com/science/article/pii/S1389128699000523},
  author = {Soumen Chakrabarti and Martin {van den Berg} and Byron Dom},
  publisher={Elsevier}
}

@article{shrivastava2023efficient,
  title={An efficient focused crawler using {LSTM-CNN} based deep learning},
  author={Shrivastava, Gourav Kumar and Pateriya, Rajesh Kumar and Kaushik, Praveen},
  journal={International Journal of System Assurance Engineering and Management},
  volume={14},
  number={1},
  pages={391--407},
  year={2023},
  publisher={Springer}
}

@inproceedings{han2018focused,
    author = {Han, Miyoung and Wuillemin, Pierre-Henri and Senellart, Pierre},
    title = {Focused Crawling Through Reinforcement Learning},
    year = {2018},
    isbn = {978-3-319-91661-3},
    publisher = {Springer-Verlag},
    url = {https://doi.org/10.1007/978-3-319-91662-0\_20},
    doi = {10.1007/978-3-319-91662-0_20},
    booktitle = {Web Engineering: 18th International Conference (ICWE 2018)},
    pages = {261--278},
    numpages = {18},
    address = {C\'{a}ceres, Spain}
}

@inproceedings{buck2016quick,
    title = "Quick and Reliable Document Alignment via {TF}/{IDF}-weighted Cosine Distance",
    author = "Buck, Christian and Koehn, Philipp",
    booktitle = "Proceedings of the First Conference on Machine Translation: Volume 2, Shared Task Papers",
    month = aug,
    year = "2016",
    address = "Berlin, Germany",
    publisher = "Association for Computational Linguistics",
    url = "https://aclanthology.org/W16-2365",
    doi = "10.18653/v1/W16-2365",
    pages = "672--678",
}

@inproceedings{guo2019hierarchical,
    title = "Hierarchical Document Encoder for Parallel Corpus Mining",
    author = "Guo, Mandy  and
      Yang, Yinfei  and
      Stevens, Keith  and
      Cer, Daniel  and
      Ge, Heming  and
      Sung, Yun-hsuan  and
      Strope, Brian  and
      Kurzweil, Ray",
    editor = "Bojar, Ond{\v{r}}ej  and
      Chatterjee, Rajen  and
      Federmann, Christian  and
      Fishel, Mark  and
      Graham, Yvette  and
      Haddow, Barry  and
      Huck, Matthias  and
      Yepes, Antonio Jimeno  and
      Koehn, Philipp  and
      Martins, Andr{\'e}  and
      Monz, Christof  and
      Negri, Matteo  and
      N{\'e}v{\'e}ol, Aur{\'e}lie  and
      Neves, Mariana  and
      Post, Matt  and
      Turchi, Marco  and
      Verspoor, Karin",
    booktitle = "Proceedings of the Fourth Conference on Machine Translation (Volume 1: Research Papers)",
    month = aug,
    year = "2019",
    address = "Florence, Italy",
    publisher = "Association for Computational Linguistics",
    url = "https://aclanthology.org/W19-5207",
    doi = "10.18653/v1/W19-5207",
    pages = "64--72",
}

@inproceedings{thompson2020exploiting,
    title = "Exploiting Sentence Order in Document Alignment",
    author = "Thompson, Brian  and
      Koehn, Philipp",
    editor = "Webber, Bonnie  and
      Cohn, Trevor  and
      He, Yulan  and
      Liu, Yang",
    booktitle = "Proceedings of the 2020 Conference on Empirical Methods in Natural Language Processing (EMNLP)",
    month = nov,
    year = "2020",
    address = "Online",
    publisher = "Association for Computational Linguistics",
    url = "https://aclanthology.org/2020.emnlp-main.483",
    doi = "10.18653/v1/2020.emnlp-main.483",
    pages = "5997--6007"
}

@inproceedings{buck2016findings,
    title = "Findings of the {WMT} 2016 Bilingual Document Alignment Shared Task",
    author = "Buck, Christian and Koehn, Philipp",
    editor = {Bojar, Ond{\v{r}}ej  and
      Buck, Christian  and
      Chatterjee, Rajen  and
      Federmann, Christian  and
      Guillou, Liane  and
      Haddow, Barry  and
      Huck, Matthias  and
      Yepes, Antonio Jimeno  and
      N{\'e}v{\'e}ol, Aur{\'e}lie  and
      Neves, Mariana  and
      Pecina, Pavel  and
      Popel, Martin  and
      Koehn, Philipp  and
      Monz, Christof  and
      Negri, Matteo  and
      Post, Matt  and
      Specia, Lucia  and
      Verspoor, Karin  and
      Tiedemann, J{\"o}rg  and
      Turchi, Marco},
    booktitle = "Proceedings of the First Conference on Machine Translation: Volume 2, Shared Task Papers",
    month = aug,
    year = "2016",
    address = "Berlin, Germany",
    publisher = "Association for Computational Linguistics",
    url = "https://aclanthology.org/W16-2347",
    doi = "10.18653/v1/W16-2347",
    pages = "554--563",
}

@inproceedings{dara2016yoda,
    title = "{YODA} System for {WMT}16 Shared Task: Bilingual Document Alignment",
    author = "Dara, Aswarth Abhilash  and Lin, Yiu-Chang",
    booktitle = "Proceedings of the First Conference on Machine Translation: Volume 2, Shared Task Papers",
    month = aug,
    year = "2016",
    address = "Berlin, Germany",
    publisher = "Association for Computational Linguistics",
    url = "https://aclanthology.org/W16-2366",
    doi = "10.18653/v1/W16-2366",
    pages = "679--684",
}

@inproceedings{el2020ccaligned,
    title = {{CCA}ligned: A Massive Collection of Cross-Lingual Web-Document Pairs},
    author = "El-Kishky, Ahmed  and
      Chaudhary, Vishrav  and
      Guzm{\'a}n, Francisco  and
      Koehn, Philipp",
    editor = "Webber, Bonnie  and
      Cohn, Trevor  and
      He, Yulan  and
      Liu, Yang",
    booktitle = "Proceedings of the 2020 Conference on Empirical Methods in Natural Language Processing (EMNLP)",
    month = nov,
    year = "2020",
    address = "Online",
    publisher = "Association for Computational Linguistics",
    url = "https://aclanthology.org/2020.emnlp-main.480",
    doi = "10.18653/v1/2020.emnlp-main.480",
    pages = "5960--5969",
}

@techreport{macocu-v2-evaluation-techreport,
    author = "van Noord, Rik  and
      Chichirau, Malina and
      Espl{\`a}-Gomis, Miquel  and
      Kuzman, Taja  and
      Ljube{\v{s}}i{\'c}, Nikola  and
      Ram{\'\i}rez-S{\'a}nchez, Gema  and
      Rupnik, Peter  and
      Toral, Antonio",
    title = "Evaluation of data release 2",
    year = "2023",
    institution = "MaCoCu Action",
    url = "https://macocu.eu/static/media/second-report.453a82100b1ec3647012.pdf",
}

@inproceedings{lui2014accurate,
    title = "Accurate Language Identification of {T}witter Messages",
    author = "Lui, Marco  and
      Baldwin, Timothy",
    editor = "Farzindar, Atefeh  and
      Inkpen, Diana  and
      Gamon, Michael  and
      Nagarajan, Meena",
    booktitle = "Proceedings of the 5th Workshop on Language Analysis for Social Media ({LASM})",
    month = apr,
    year = "2014",
    address = "Gothenburg, Sweden",
    publisher = "Association for Computational Linguistics",
    url = "https://aclanthology.org/W14-1303",
    doi = "10.3115/v1/W14-1303",
    pages = "17--25",
}

@inproceedings{wolf-etal-2020-transformers,
    title = "Transformers: State-of-the-Art Natural Language Processing",
    author = "Thomas Wolf and Lysandre Debut and Victor Sanh and Julien Chaumond and Clement Delangue and Anthony Moi and Pierric Cistac and Tim Rault and R{\'e}mi Louf and Morgan Funtowicz and Joe Davison and Sam Shleifer and Patrick von Platen and Clara Ma and Yacine Jernite and Julien Plu and Canwen Xu and Teven Le Scao and Sylvain Gugger and Mariama Drame and Quentin Lhoest and Alexander M. Rush",
    booktitle = "Proceedings of the 2020 Conference on Empirical Methods in Natural Language Processing: System Demonstrations",
    month = oct,
    year = "2020",
    address = "Online",
    publisher = "Association for Computational Linguistics",
    url = "https://www.aclweb.org/anthology/2020.emnlp-demos.6",
    pages = "38--45"
}

@inproceedings{loshchilov2017decoupled,
  title={Decoupled weight decay regularization},
  author={Loshchilov, Ilya and Hutter, Frank},
  booktitle = "Proceedings of the Seventh International Conference on Learning Representations (ICLR 2019)",
  address = "New Orleans, USA",
  pages = "1--8",
  year={2019}
}

@inproceedings{devlin-etal-2019-bert,
    title = "{BERT}: Pre-training of Deep Bidirectional Transformers for Language Understanding",
    author = "Devlin, Jacob  and
      Chang, Ming-Wei  and
      Lee, Kenton  and
      Toutanova, Kristina",
    booktitle = "Proceedings of the 2019 Conference of the North {A}merican Chapter of the Association for Computational Linguistics: Human Language Technologies, Volume 1 (Long and Short Papers)",
    month = jun,
    year = "2019",
    address = "Minneapolis, Minnesota",
    publisher = "Association for Computational Linguistics",
    url = "https://aclanthology.org/N19-1423",
    doi = "10.18653/v1/N19-1423",
    pages = "4171--4186"
}

@inproceedings{zhang2020revisiting,
  title={Revisiting Few-sample BERT Fine-tuning},
  author={Zhang, Tianyi and Wu, Felix and Katiyar, Arzoo and Weinberger, Kilian Q and Artzi, Yoav},
  booktitle = "Proceedings of the Ninth International Conference on Learning Representations (ICLR 2021)",
  address = "Virtual",
  pages = "1--22",
  year={2021}
}

@inproceedings{paszke2019pytorch,
  title={{PyTorch}: An imperative style, high-performance deep learning library},
  author={Paszke, Adam and Gross, Sam and Massa, Francisco and Lerer, Adam and Bradbury, James and Chanan, Gregory and Killeen, Trevor and Lin, Zeming and Gimelshein, Natalia and Antiga, Luca and Desmaison, Alban and Kopf, Andreas and Yang, Edward and DeVito, Zachary and Raison, Martin and Tejani, Alykhan and Chilamkurthy, Sasank and Steiner, Benoit Fang, Lu and Bai, Junjie and Chintala, Soumith},
  booktitle={33rd Conference on Neural Information Processing Systems (NeurIPS 2019)},
  address = {Vancouver, Canada},
  pages = "1--12",
  year={2019}
}

@inproceedings{schwenk2021ccmatrix,
    title = "{CCM}atrix: Mining Billions of High-Quality Parallel Sentences on the Web",
    author = "Schwenk, Holger  and
      Wenzek, Guillaume  and
      Edunov, Sergey  and
      Grave, Edouard  and
      Joulin, Armand  and
      Fan, Angela",
    editor = "Zong, Chengqing  and
      Xia, Fei  and
      Li, Wenjie  and
      Navigli, Roberto",
    booktitle = "Proceedings of the 59th Annual Meeting of the Association for Computational Linguistics and the 11th International Joint Conference on Natural Language Processing (Volume 1: Long Papers)",
    month = aug,
    year = "2021",
    address = "Online",
    publisher = "Association for Computational Linguistics",
    url = "https://aclanthology.org/2021.acl-long.507",
    doi = "10.18653/v1/2021.acl-long.507",
    pages = "6490--6500"
}

@inproceedings{agarwal2014focused,
author = {Agarwal, Swati and Sureka, Ashish},
title = {A Focused Crawler for Mining Hate and Extremism Promoting Videos on YouTube.},
publisher = {Association for Computing Machinery},
address = {Santiago, Chile},
url = {https://doi.org/10.1145/2631775.2631776},
doi = {10.1145/2631775.2631776},
booktitle = {Proceedings of the 25th ACM Conference on Hypertext and Social Media},
year = "2014",
pages = {294--296}
}

@inproceedings{abbasi2013crawling,
  title={Crawling credible online medical sentiments for social intelligence},
  author={Abbasi, Ahmed and Fu, Tianjun and Zeng, Daniel and Adjeroh, Donald},
  booktitle={2013 International Conference on Social Computing},
  pages={254--263},
  year={2013},
  organization={IEEE}
}

@article{espla2010combining,
  title={Combining Content-Based and {URL}-Based Heuristics to Harvest Aligned Bitexts from Multilingual Sites with Bitextor},
  author={Espl{\`a}-Gomis, Miquel and Forcada, Mikel L.},
  journal={The Prague Bulletin of Mathematical Linguistics},
  number={93},
  publisher={Charles University, Czech Republic Faculty of Mathematics and Physics},
  pages={77--86},
  year={2010}
}

@article{hernandez2016cala,
  title={{CALA}: {ClAssifying} Links Automatically based on their {URL}},
  author={Hern{\'a}ndez, Inma and Rivero, Carlos R and Ruiz, David and Corchuelo, Rafael},
  journal={Journal of Systems and Software},
  volume={115},
  pages={130--143},
  year={2016},
  url = {https://www.sciencedirect.com/science/article/pii/S016412121600042X},
  publisher={Elsevier}
}

@inproceedings{rajalakshmi2013web,
  title={Web page classification using n-gram based {URL} features},
  author={Rajalakshmi, Ratnavel and Aravindan, Chandrabose},
  booktitle={2013 Fifth International Conference on Advanced Computing ({ICoAC})},
  pages={15--21},
  year={2013},
  address={Chennai, India},
  organization={IEEE}
}

@inproceedings{chen2000parallel,
author = {Chen, Jiang and Nie, Jian-Yun},
title = {Parallel Web Text Mining for Cross-Language {IR}},
year = {2000},
address = {Paris, France},
booktitle = {Content-Based Multimedia Information Access - Volume 1},
pages = {62--77}
}

@article{resnik2003web,
    author = {Resnik, Philip and Smith, Noah A.},
    title = {The Web as a Parallel Corpus},
    journal = {Computational Linguistics},
    volume = {29},
    number = {3},
    pages = {349--380},
    year = {2003},
    month = {09},
    issn = {0891-2017},
    doi = {10.1162/089120103322711578},
    url = {https://doi.org/10.1162/089120103322711578},
    eprint = {https://direct.mit.edu/coli/article-pdf/29/3/349/1798141/089120103322711578.pdf},
    publisher={MIT Press}
}

@inproceedings{barbosa2011crawling,
    title = "Crawling Back and Forth: Using Back and Out Links to Locate Bilingual Sites",
    author = "Barbosa, Luciano  and
      Bangalore, Srinivas  and
      Rangarajan Sridhar, Vivek Kumar",
    editor = "Wang, Haifeng  and
      Yarowsky, David",
    booktitle = "Proceedings of 5th International Joint Conference on Natural Language Processing",
    month = nov,
    year = "2011",
    address = "Chiang Mai, Thailand",
    publisher = "Asian Federation of Natural Language Processing",
    url = "https://aclanthology.org/I11-1048",
    pages = "429--437",
}

@inproceedings{zhang2013finding,
    title = "Finding More Bilingual Webpages with High Credibility via Link Analysis",
    author = "Zhang, Chengzhi  and
      Yao, Xuchen  and
      Kit, Chunyu",
    editor = "Sharoff, Serge  and
      Zweigenbaum, Pierre  and
      Rapp, Reinhard",
    booktitle = "Proceedings of the Sixth Workshop on Building and Using Comparable Corpora",
    month = aug,
    year = "2013",
    address = "Sofia, Bulgaria",
    publisher = "Association for Computational Linguistics",
    url = "https://aclanthology.org/W13-2517",
    pages = "138--143",
}

@inproceedings{reid-artetxe-2022-paradise,
    title = "{PARADISE}: Exploiting Parallel Data for Multilingual Sequence-to-Sequence Pretraining",
    author = "Reid, Machel  and
      Artetxe, Mikel",
    editor = "Carpuat, Marine  and
      de Marneffe, Marie-Catherine  and
      Meza Ruiz, Ivan Vladimir",
    booktitle = "Proceedings of the 2022 Conference of the North American Chapter of the Association for Computational Linguistics: Human Language Technologies",
    month = jul,
    year = "2022",
    address = "Seattle, United States",
    publisher = "Association for Computational Linguistics",
    url = "https://aclanthology.org/2022.naacl-main.58",
    doi = "10.18653/v1/2022.naacl-main.58",
    pages = "800--810"
}

@misc{cldeval14,
author = {CLD2Owners},
title = {Evaluate CLD2 20140122 1024k},
howpublished = {\url{https://github.com/CLD2Owners/cld2/blob/master/docs/evaluate_cld2_large_20140122.txt}},
note = {Accessed: February 5, 2024},
year = {2014}
}

@inproceedings{kale-etal-2021-nmt5,
    title = "nm{T}5 - Is parallel data still relevant for pre-training massively multilingual language models?",
    author = "Kale, Mihir  and
      Siddhant, Aditya  and
      Al-Rfou, Rami  and
      Xue, Linting  and
      Constant, Noah  and
      Johnson, Melvin",
    editor = "Zong, Chengqing  and
      Xia, Fei  and
      Li, Wenjie  and
      Navigli, Roberto",
    booktitle = "Proceedings of the 59th Annual Meeting of the Association for Computational Linguistics and the 11th International Joint Conference on Natural Language Processing (Volume 2: Short Papers)",
    month = aug,
    year = "2021",
    address = "Online",
    publisher = "Association for Computational Linguistics",
    url = "https://aclanthology.org/2021.acl-short.87",
    doi = "10.18653/v1/2021.acl-short.87",
    pages = "683--691",
}

@article{smith2022using,
  title={Using deepspeed and megatron to train megatron-turing nlg 530b, a large-scale generative language model},
  author={Smith, Shaden and Patwary, Mostofa and Norick, Brandon and LeGresley, Patrick and Rajbhandari, Samyam and Casper, Jared and Liu, Zhun and Prabhumoye, Shrimai and Zerveas, George and Korthikanti, Vijay and others},
  journal={arXiv preprint arXiv:2201.11990},
  month=feb,
  year={2022},
  url="https://arxiv.org/abs/2201.11990",
  doi="10.48550/arXiv.2201.11990",
}

@inproceedings{xue-etal-2021-mt5,
    title = "m{T}5: A Massively Multilingual Pre-trained Text-to-Text Transformer",
    author = "Xue, Linting  and
      Constant, Noah  and
      Roberts, Adam  and
      Kale, Mihir  and
      Al-Rfou, Rami  and
      Siddhant, Aditya  and
      Barua, Aditya  and
      Raffel, Colin",
    editor = "Toutanova, Kristina  and
      Rumshisky, Anna  and
      Zettlemoyer, Luke  and
      Hakkani-Tur, Dilek  and
      Beltagy, Iz  and
      Bethard, Steven  and
      Cotterell, Ryan  and
      Chakraborty, Tanmoy  and
      Zhou, Yichao",
    booktitle = "Proceedings of the 2021 Conference of the North American Chapter of the Association for Computational Linguistics: Human Language Technologies",
    month = jun,
    year = "2021",
    address = "Online",
    publisher = "Association for Computational Linguistics",
    url = "https://aclanthology.org/2021.naacl-main.41",
    doi = "10.18653/v1/2021.naacl-main.41",
    pages = "483--498",
}

@inproceedings{joulin-etal-2017-bag,
    title = "Bag of Tricks for Efficient Text Classification",
    author = "Joulin, Armand  and
      Grave, Edouard  and
      Bojanowski, Piotr  and
      Mikolov, Tomas",
    editor = "Lapata, Mirella  and
      Blunsom, Phil  and
      Koller, Alexander",
    booktitle = "Proceedings of the 15th Conference of the {E}uropean Chapter of the Association for Computational Linguistics: Volume 2, Short Papers",
    month = apr,
    year = "2017",
    address = "Valencia, Spain",
    publisher = "Association for Computational Linguistics",
    url = "https://aclanthology.org/E17-2068/",
    pages = "427--431"
}

@misc{fasttext-langid,
  title        = {Language Identification with fastText},
  author       = {{Facebook AI Research}},
  howpublished = {\url{https://fasttext.cc/docs/en/language-identification.html}},
  note         = {Accessed: December 16, 2025},
  year         = {2017} % https://github.com/facebookresearch/fastText/issues/203#issuecomment-335491163
}

@article{nllb,
  title={No Language Left Behind: Scaling Human-Centered Machine Translation},
  author={{NLLB Team} and Costa-juss{\`a}, Marta R. and Cross, James and {\c{C}}elebi, Onur and Elbayad, Maha and Heafield, Kenneth and Heffernan, Kevin and Kalbassi, Elahe and Lam, Janice and Licht, Daniel and Maillard, Jean and Sun, Anna and Wang, Skyler and Wenzek, Guillaume and Youngblood, Al and Akula, Bapi and Barrault, Loic and Gonzalez Mejia, Gabriel and Hansanti, Prangthip and Hoffman, John and Jarrett, Semarley and Sadagopan, Kaushik Ram and Rowe, Dirk and Spruit, Shannon and Tran, Chau and Andrews, Pierre and Ayan, Necip Fazil and Bhosale, Shruti and Edunov, Sergey and Fan, Angela and Gao, Cynthia and Goswami, Vedanuj and Guzm{\'a}n, Francisco and Koehn, Philipp and Mourachko, Alexandre and Ropers, Christophe and Saleem, Safiyyah and Schwenk, Holger and Wang, Jeff},
  journal={arXiv preprint arXiv:2207.04672},
  year={2022},
  doi={10.48550/arXiv.2207.04672},
}

\appendix

\section{Synthetic Negative Samples for Inferring Parallelness from URLs}\label{ap:synthetic}
This Appendix describes the evaluation of the different strategies described in Section~\ref{se:classifier-synthetic-negative-samples} to produce synthetic negative samples consisting of pairs of URLs corresponding to non-parallel documents. We assess all combinations of these strategies 
on the WMT16 dataset, for which negative samples can be straightforwardly obtained from the HTML content. 

Our approach involves using only the WMT16 training split to generate negative samples using the combination of strategies under evaluation, and adopting a 10-fold cross-validation strategy. In each cross-validation iteration, we extract negative samples from the HTML documents in the test fold, and use the synthetic negative samples under evaluation on the remaining folds. Subsequently, we employ the positive samples and synthetic negative samples from the training folds to train a classifier, which is then evaluated on the data of the test fold.

Table~\ref{table:eval-negative-samples} shows the results obtained for all these possible combinations. These results confirm that the most effective combination includes \textit{random match} (both monolingual and bilingual), \textit{maximize Jaccard similarity} (both monolingual and bilingual), and \textit{remove random token} (bilingual only), with a macro F1 of $95.74\%$ (positive class F1: $92.00\%$; negative class F1: $99.49\%$).

\begin{table*}[t]
\begin{threeparttable}
\begin{tabular}[t]{lrrr||lrrr}
 \toprule

 \headrow Methods\tnote{a} & Pos. F1 & Neg. F1 & Macro F1 & Methods\tnote{a} & Pos. F1 & Neg. F1 & Macro F1 \\
 \midrule
1 & 75.31\% & 98.02\% & 86.66\% & 2+3+5 & 88.45\% & 99.24\% & 93.85\% \\
2 & 85.49\% & 99.03\% & 92.26\% & 2+3+6 & 84.88\% & 98.99\% & 91.93\% \\
3 & 27.87\% & 81.93\% & 54.90\% & 2+4+5 & 88.63\% & 99.25\% & 93.94\% \\
4 & 79.73\% & 98.37\% & 89.05\% & 2+4+6 & 88.78\% & 99.27\% & 94.02\% \\
5 & 58.43\% & 95.68\% & 77.06\% & 2+5+6 & 91.22\% & 99.44\% & 95.33\% \\
6 & 20.50\% & 67.82\% & 44.16\% & 3+4+5 & 90.42\% & 99.35\% & 94.89\% \\
1+2 & 84.07\% & 98.89\% & 91.48\% & 3+4+6 & 86.58\% & 99.06\% & 92.82\% \\
1+3 & 77.33\% & 98.24\% & 87.79\% & 3+5+6 & 78.82\% & 98.38\% & 88.60\% \\
1+4 & 84.89\% & 98.93\% & 91.91\% & 4+5+6 & 90.87\% & 99.40\% & 95.14\% \\
1+5 & 88.47\% & 99.24\% & 93.85\% & 1+2+3+4 & 89.21\% & 99.29\% & 94.25\% \\
1+6 & 74.49\% & 97.94\% & 86.22\% & 1+2+3+5 & 89.40\% & 99.32\% & 94.36\% \\
2+3 & 87.60\% & 99.16\% & 93.38\% & 1+2+3+6 & 89.22\% & 99.31\% & 94.27\% \\
2+4 & 88.56\% & 99.25\% & 93.91\% & 1+2+4+5 & 88.03\% & 99.21\% & 93.62\% \\
2+5 & 88.48\% & 99.24\% & 93.86\% & 1+2+4+6 & 87.83\% & 99.18\% & 93.51\% \\
2+6 & 88.00\% & 99.22\% & 93.61\% & 1+2+5+6 & 90.40\% & 99.39\% & 94.90\% \\
3+4 & 88.96\% & 99.26\% & 94.11\% & 1+3+4+5 & 90.72\% & 99.39\% & 95.06\% \\
3+5 & 82.39\% & 98.74\% & 90.56\% & 1+3+4+6 & 83.84\% & 98.86\% & 91.35\% \\
3+6 & 24.87\% & 76.66\% & 50.77\% & 1+3+5+6 & 91.27\% & 99.44\% & 95.35\% \\
4+5 & 62.00\% & 96.29\% & 79.14\% & 1+4+5+6 & 91.52\% & 99.45\% & 95.48\% \\
4+6 & 69.23\% & 97.22\% & 83.22\% & 2+3+4+5 & 89.59\% & 99.32\% & 94.45\% \\
5+6 & 76.93\% & 98.18\% & 87.55\% & 2+3+4+6 & 88.17\% & 99.21\% & 93.69\% \\
1+2+3 & 88.98\% & 99.27\% & 94.12\% & 2+3+5+6 & 89.22\% & 99.30\% & 94.26\% \\
1+2+4 & 85.39\% & 98.98\% & 92.18\% & 2+4+5+6 & 90.24\% & 99.36\% & 94.80\% \\
1+2+5 & 88.41\% & 99.25\% & 93.83\% & 3+4+5+6 & 89.73\% & 99.31\% & 94.52\% \\
1+2+6 & 86.76\% & 99.12\% & 92.94\% & \textbf{1+2+3+4+5} & \textbf{92.00\%} & \textbf{99.49\%} & \textbf{95.74\%} \\
1+3+4 & 84.53\% & 98.92\% & 91.72\% & 1+2+3+4+6 & 85.15\% & 99.01\% & 92.08\% \\
1+3+5 & 90.86\% & 99.41\% & 95.14\% & 1+2+3+5+6 & 90.52\% & 99.39\% & 94.95\% \\
1+3+6 & 70.89\% & 97.48\% & 84.18\% & 1+2+4+5+6 & 87.93\% & 99.19\% & 93.56\% \\
1+4+5 & 90.14\% & 99.35\% & 94.74\% & 1+3+4+5+6 & 89.06\% & 99.29\% & 94.17\% \\
1+4+6 & 84.48\% & 98.89\% & 91.69\% & 2+3+4+5+6 & 91.15\% & 99.43\% & 95.29\% \\
1+5+6 & 89.58\% & 99.32\% & 94.45\% & 1+2+3+4+5+6 & 88.87\% & 99.27\% & 94.07\% \\
2+3+4 & 90.38\% & 99.38\% & 94.88\% & & & & \\
\bottomrule
\end{tabular}
\begin{tablenotes}[hang]
\item[a]Codification: \textit{random match bilingual} (1), \textit{maximize Jaccard similarity bilingual} (2), \textit{remove random tokens bilingual} (3), \textit{random match monolingual} (4), \textit{maximize Jaccard similarity monolingual} (5), \textit{remove random tokens monolingual} (6).
\end{tablenotes}
\caption{Results of the evaluation of all combinations of methods to generate synthetic negative samples. The best combination appears in boldface.}
\label{table:eval-negative-samples}
\end{threeparttable}
\end{table*}

\end{document}